\documentclass{article}
\usepackage{footnote}
\makesavenoteenv{figure}
\usepackage{arxiv}
\usepackage[font=small]{caption}
\usepackage[utf8]{inputenc} 
\usepackage[T1]{fontenc}    

\usepackage[dvipsnames]{xcolor}
\definecolor{kimiblue}{rgb}{0.09,0.5,0.99}
\usepackage[breaklinks,colorlinks,allcolors=kimiblue]{hyperref}

\usepackage{url}            
\usepackage{booktabs}       
\usepackage{amsfonts}       
\usepackage{nicefrac}       
\usepackage{microtype}      
\usepackage{lipsum}		
\usepackage{graphicx}
\usepackage{doi}
\usepackage{multirow}
\usepackage{multicol}
\usepackage{xcolor}
\usepackage{threeparttable}

\usepackage{mathrsfs}
\usepackage[normalem]{ulem}
\usepackage{amsmath,amsfonts,bm}
\usepackage{amssymb,amsthm}
\usepackage{enumitem}
\usepackage{adjustbox}
\usepackage{subcaption}
\usepackage{adjustbox}
\usepackage{array}
\usepackage{makecell}

\usepackage{wrapfig}
\usepackage{ulem}
\usepackage[
    backend=biber,
    citestyle=numeric,
    bibstyle=numeric,
    mincitenames=1,
    maxcitenames=1,
    maxbibnames=3,
]{biblatex}

\usepackage{tikz}
\usepackage{color}
\usepackage{threeparttable}
\usepackage{multirow}
\usepackage{multicol}
\usepackage{colortbl}
\usepackage{geometry}
\usepackage{mathtools}
\usepackage{pgfplots}
\usepackage{subcaption}
\usepackage{graphicx}
\usepackage[frozencache,cachedir=.]{minted} 

\usepackage[most]{tcolorbox}
\usepackage{algorithm}
\usepackage{algpseudocode}
\usepackage{fontawesome}

\newtheorem{proposition}{Proposition}
\usetikzlibrary{
  arrows.meta,
  positioning,
  calc,
  shapes.geometric,
  shapes.misc,
  decorations.text,
  patterns,
  patterns.meta,
  fit,
  backgrounds,
  chains,
  shadows,
  math,
  matrix,
  circuits.ee.IEC,
  decorations.pathmorphing,
  decorations.pathreplacing,
  decorations.shapes,
  decorations.pathreplacing,
  calligraphy
}

\definecolor{brickred}{HTML}{b92622}
\definecolor{midnightblue}{HTML}{005c7f}
\definecolor{limegreen}{HTML}{97c65a}
\definecolor{salmon}{HTML}{f1958d}
\definecolor{darkcyan}{HTML}{008B8B}
\definecolor{darkgrey}{rgb}{0.53,0.53,0.53}
\definecolor{mygrey}{rgb}{0.9,0.9,0.9}

\newcommand{\white}[1]{\textcolor{white}{#1}}
\newcommand{\brickred}[1]{\textcolor{brickred}{#1}}
\newcommand{\midnightblue}[1]{\textcolor{midnightblue}{#1}}

\newcommand{\citep}[1]{\parencite{#1}}
\newcommand{\github}{\raisebox{0pt}{\faGithub}}

\newcommand{\huggingface}{\raisebox{-2pt}{\includegraphics[scale=0.038]{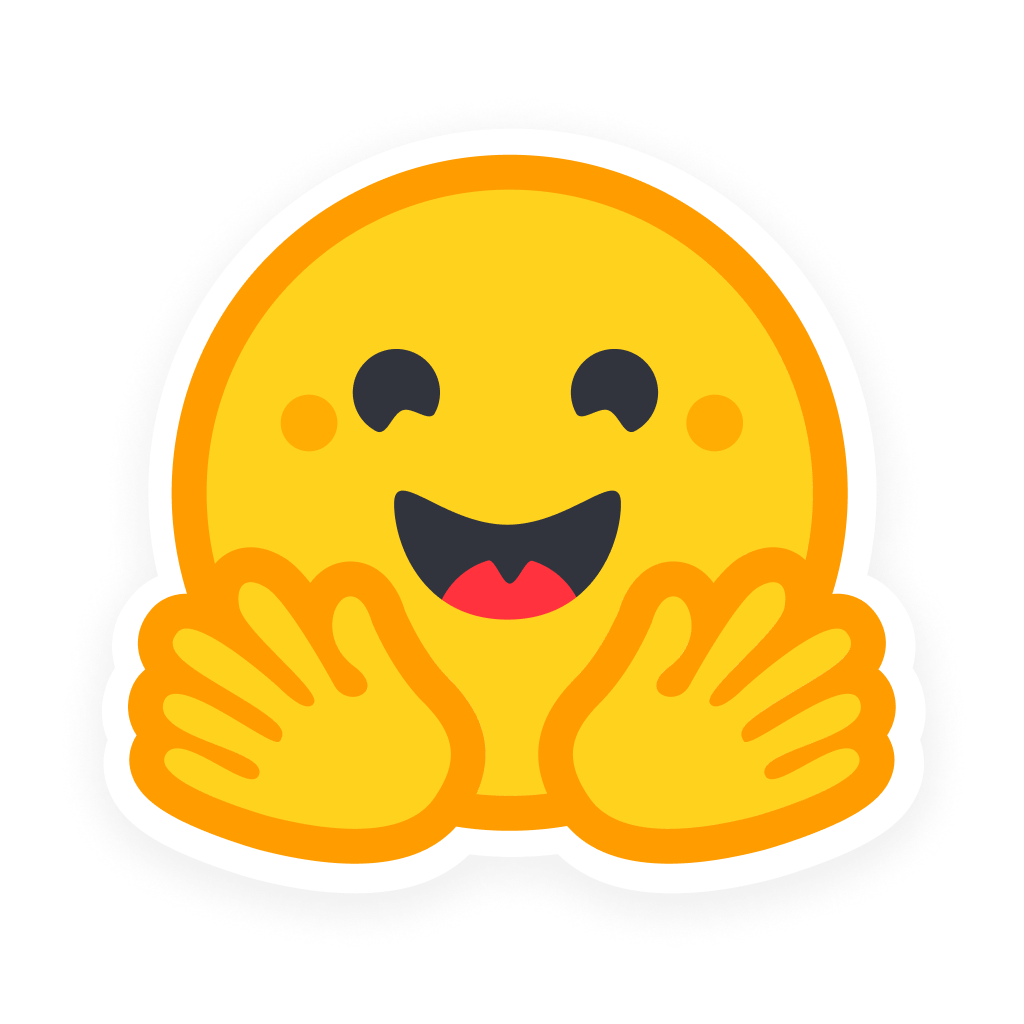}}}

\setlist[itemize,1]{leftmargin=\dimexpr 18pt}
\setlist[enumerate,1]{leftmargin=\dimexpr 18pt}

\title{
\raisebox{-0.1\height}{
\includegraphics[width=0.032\textwidth]{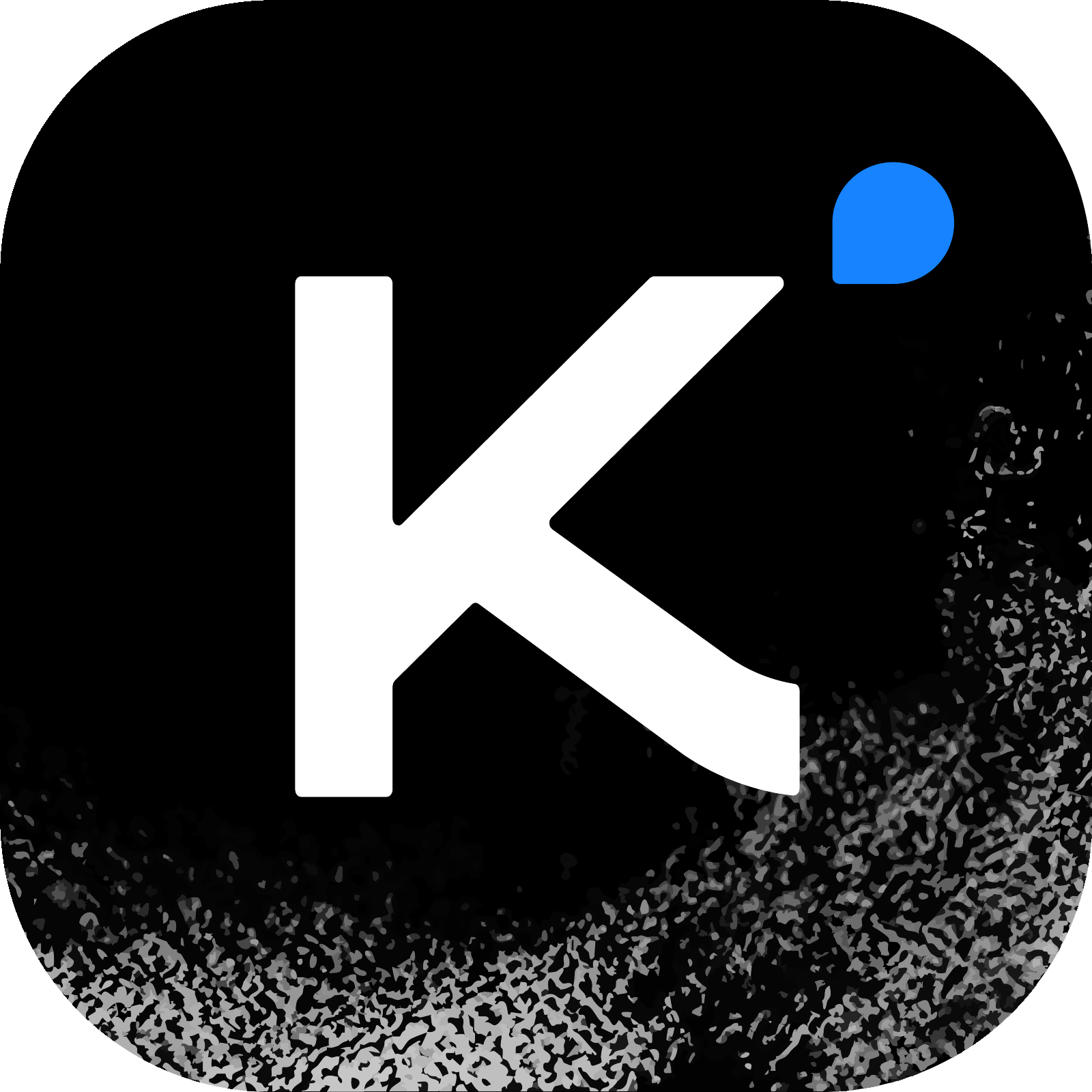}} %
Kimi Linear:\\ An Expressive, Efficient Attention Architecture
}

\author{
Kimi Team\\
\github\,\,\url{https://github.com/MoonshotAI/Kimi-Linear}\\
}

\date{}

\renewcommand{\headeright}{Technical Report}
\renewcommand{\undertitle}{Technical Report of Kimi Linear}
\renewcommand{\shorttitle}{\raisebox{-0.12\height}{\includegraphics[width=0.02\textwidth]{figures/logo.pdf}}
Kimi Linear: An Expressive, Efficient Attention Architecture}

\begin{document}
\maketitle

\vspace{-15pt}
\begin{abstract}
We introduce Kimi Linear, a hybrid linear attention architecture that, for the first time, outperforms full attention under fair comparisons across various scenarios—including short-context, long-context, and reinforcement learning (RL) scaling regimes. At its core lies Kimi Delta Attention (KDA), an expressive linear attention module that extends Gated DeltaNet \citep{yang-2025-gdn} with a finer-grained gating mechanism, enabling more effective use of limited finite-state RNN memory. Our bespoke chunkwise algorithm achieves high hardware efficiency through a specialized variant of the \emph{Diagonal-Plus-Low-Rank} (DPLR) transition matrices, which substantially reduces computation compared to the general DPLR formulation while remaining more consistent with the classical delta rule.

We pretrain a Kimi Linear model with 3B activated parameters and 48B total parameters, based on a layerwise hybrid of KDA and Multi-Head Latent Attention (MLA). Our experiments show that with an identical training recipe, Kimi Linear outperforms full MLA with a sizeable margin across all evaluated tasks, while reducing KV cache usage by up to 75\% and achieving up to $6\times$ decoding throughput for a 1M context. These results demonstrate that Kimi Linear can be a drop-in replacement for full attention architectures with superior performance and efficiency, including tasks with longer input and output lengths.

To support further research, we open-source the KDA kernel and vLLM implementations \footnote{\github\,\,\url{https://github.com/fla-org/flash-linear-attention/tree/main/fla/ops/kda}}, and release the pre-trained and instruction-tuned model checkpoints. \footnote{\huggingface\,\,\url{https://huggingface.co/moonshotai/Kimi-Linear-48B-A3B-Instruct}}

\end{abstract}

\input{figures/mainfig}

\section{Introduction}

As large language models (LLMs) evolve into increasingly capable agents~\citep{kimi2025k2}, the computational demands of inference—particularly in long-horizon and reinforcement learning (RL) settings—are becoming a central bottleneck. This shift toward \textit{RL test-time scaling}~\citep{kimiteam2025kimik15scalingreinforcement,guo2025deepseek,qu2025survey,plaat2024reasoning,lai2025survey}, where models must process extended trajectories, tool-use interactions, and complex decision spaces at inference time, exposes fundamental inefficiencies in standard attention mechanisms. In particular, the quadratic time complexity and the linearly growing key–value (KV) cache of softmax attention introduce substantial computational and memory overheads, hindering throughput, context-length scaling, and real-time interactivity.

Linear attention~\citep{katharopoulos-2020-transformers} offers a principled approach to reducing computational complexity but has historically underperformed softmax attention in language modeling—even for short sequences—due to limited expressivity. Recent advances have significantly narrowed this gap, primarily through two innovations: gating or decay mechanisms~\citep{sun-2023-retnet,mamba2,yang-etal-2024-gla} and the delta rule~\citep{schlag-2021-deltanet,yang-2024-parallelizing,yang-2025-gdn,peng-2025-rwkv7}. Together, these developments have pushed linear attention closer to softmax-level quality on moderate-length sequences. Nevertheless, purely linear structure remain fundamentally constrained by the finite-state capacity, making long-sequence modeling and in-context retrieval theoretically challenging~\citep{wen2024rnns,arora2024simple,jelassi-2024-repeat}.  

Hybrid architectures that combine softmax and linear attention—using a few global-attention layers alongside predominantly faster linear layers—have thus emerged as a practical compromise between quality and efficiency~\citep{lieber2024jamba,mamba2hybrid,minimax2025minimax01,blakeman2025nemotron,gu2025jetnemotronefficientlanguagemodel,qwen3next2025}. However, previous hybrid models often operated at limited scale or lacked comprehensive evaluation across diverse benchmarks. The core challenge remains: to develop an attention architecture that matches or surpasses full attention in quality while achieving substantial efficiency gains in both speed and memory—an essential step toward enabling the next generation of agentic, decoding-heavy LLMs.  

In this work, we present \textbf{Kimi Linear}, a hybrid linear attention architecture designed to meet the efficiency demands of agentic intelligence and test-time scaling without compromising quality. At its core lies \textbf{Kimi Delta Attention (KDA)}, a hardware-efficient linear attention module that extends Gated DeltaNet~\citep{yang-2025-gdn} with a finer-grained gating mechanism. While GDN, similar to Mamba2~\citep{mamba2}, employs a coarse head-wise forget gate, KDA introduces a channel-wise variant in which each feature dimension maintains an independent forgetting rate, akin to Gated Linear Attention (GLA)~\citep{yang-etal-2024-gla}. This fine-grained design enables more precise regulation of the finite-state RNN memory, unlocking the potential of RNN-style models within hybrid architectures.  

Crucially, KDA parameterizes its transition dynamics with a specialized variant of the \emph{Diagonal-Plus-Low-Rank} (DPLR) matrices \cite{gu-2022-efficiently,peng-2025-rwkv7}, enabling a bespoke chunkwise-parallel algorithm that substantially reduces computation relative to general DPLR formulations while remaining consistent with the classical delta rule.  

Kimi Linear interleaves KDA with periodic full attention layers in a uniform 3:1 ratio. This hybrid structure reduces memory and KV-cache usage by up to 75\% during long-sequence generation while preserving global information flow via the full attention layers. Through matched-scale pretraining and evaluation, we show that Kimi Linear consistently matches or outperforms strong full-attention baselines across short-context, long-context, and RL-style post-training tasks—while achieving up to 6$\times$ higher decoding throughput at 1M context length.

To facilitate further research, we release open-source KDA kernels with vLLM integration, as well as pre-trained and instruction-tuned checkpoints. These components are drop-in compatible with existing full-attention pipelines, requiring no modification to caching or scheduling interfaces, thereby facilitating research on hybrid architectures.

\paragraph{Contributions}
\begin{itemize}[leftmargin=12pt]
    \item \textbf{Kimi Delta Attention (KDA):} a linear attention mechanism that refines the gated delta rule with improved recurrent memory management and hardware efficiency.
    \item \textbf{The Kimi Linear architecture:} a hybrid design adopting a 3:1 KDA-to-global attention ratio, reducing memory footprint while surpassing full-attention quality.
    \item \textbf{Fair empirical validation at scale:} through 1.4T token training runs, Kimi Linear outperforms full attention and other baselines in short/long context and RL-style evaluations, with full release of kernels, vLLM integration, and checkpoints.
\end{itemize}

\section{Preliminary}

In this section, we introduce the technical background related to our proposed Kimi Delta Attention.

\subsection{Notation}
In this paper, we define $\square_t \in \mathbb{R}^{d_k}$ or $\mathbb{R}^{d_v}$, $\operatorname{s.t.}, \square\in \{\bm{q,k,v,o,u,w}\}$ denotes a $t$-th corresponding column vector, and $\mathbf{S}_t \in \mathbb{R}^{d_k \times d_v}$ represents the matrix-form memory state. $\mathbf{M}$ and $\mathbf{M}^{-}$ denote lower-triangular masks with and without diagonal elements, respectively; for convenience, we also write them as $\operatorname{Tril}$ and $\operatorname{StrictTril}$.

\paragraph{Chunk-wise Formulation}
Suppose the sequence is split into $L/C$ chunks where each chunk is of length $C$. We define $\square_{[t]} \in \mathbb{R}^{C\times d}$ for $\square \in \{\mathbf{Q}, \mathbf{K}, \mathbf{V}, \mathbf{O}, \mathbf{U}, \mathbf{W} \}$ are matrices that stack the vectors within the $t$-th chunk, and $\square_{[t]}^r=\square_{tC+r}$ is the $r$-th element of the chunk. Note that $t\in [0, L/C), r \in [1, C]$. State matrices are also re-indexed such that $\bm{S}^i_{[t]} = \bm{S}_{tC+i}$. Additionally, $\bm{S}_{[t]}:=\bm{S}_{[t]}^0 = \bm{S}_{[t-1]}^C$, i.e., the initial state of a chunk is the last state of the previous chunk. 

\paragraph{Decay Formulation}
We define the cumulative decay $\brickred{\gamma_{[t]}^{i\rightarrow j}} := \prod_{k=i}^j \brickred{\alpha_{[t]}^k}$, and abbreviate $\brickred{{\gamma}_{[t]}^{1\rightarrow r}}$ as $\brickred{{\gamma}_{[t]}^{r}}$. 
Additionally, $\brickred{\mathcal{A}_{[t]}} :=\brickred{\mathcal{A}_{[t]}^{i/j}}\in\mathbb{R}^{C\times C}$ is the matrix with elements $\brickred{\gamma_{[t]}^i/\gamma_{[t]}^j}$. 
$\brickred{\operatorname{Diag}\left(\bm{\alpha}_t\right)}$ denotes the fine-grained decay, $\brickred{\operatorname{Diag}\left(\boldsymbol{\gamma}_{[t]}^{i\rightarrow j}\right)} := \prod_{k=i}^j \brickred{\operatorname{Diag}\left(\boldsymbol{\alpha}_{[t]}^k\right)}$, and $\brickred{\bm{\Gamma}_{[t]}^{i\rightarrow j}}\in\mathbb{R}^{C\times d_k}$ is the matrix stack from $\brickred{\boldsymbol{\gamma}_{[t]}^i}$ to $\brickred{\boldsymbol{\gamma}_{[t]}^j}$.

\subsection{Linear Attention and the Gated Delta Rule}

\paragraph{Linear Attention as Online Learning.}
Linear attention~\citep{katharopoulos-2020-transformers} maintains a matrix-valued recurrent state that accumulates key--value associations:
\[
\mathbf{S}_t = \mathbf{S}_{t-1} + \bm{k}_t \bm{v}_t^\top, \qquad 
\bm{o}_t = \mathbf{S}_t^\top \bm{q}_t .
\]
From the fast-weight perspective~\citep{schlag-2021-deltanet,schlag-2021-learning}, $\mathbf{S}_t$ serves as an associative memory storing transient mappings from keys to values.  
This update can be viewed as performing gradient \emph{descent} on the unbounded correlation objective
\[
\mathcal{L}_t(\mathbf{S}) = -\langle \mathbf{S}^\top \bm{k}_t, \bm{v}_t \rangle ,
\]
which continually reinforces recent key--value pairs without any forgetting.  
However, such an objective provides no criterion for which memories to erase, and the accumulated state grows unbounded, leading to interference over long contexts.  

\paragraph{DeltaNet: Online Gradient Descent on Reconstruction Loss.}
DeltaNet~\citep{schlag-2021-deltanet} reinterprets this recurrence as online gradient \emph{descent} on a reconstruction objective:
\[
\mathcal{L}_t(\mathbf{S}) = \tfrac{1}{2}\|\mathbf{S}^\top\bm{k}_t - \bm{v}_t\|^2 .
\]
Taking a gradient step with learning rate $\beta_t$ gives
\[
\mathbf{S}_t
= \mathbf{S}_{t-1} - \beta_t \nabla_\mathbf{S}\mathcal{L}_t(\mathbf{S}_{t-1})
=(\mathbf{I}-\beta_t\bm{k}_t\bm{k}_t^\top)  \mathbf{S}_{t-1}
  + \beta_t\bm{k}_t\bm{v}_t^\top .
\]
This rule---the classical \emph{delta rule}---treats $\mathbf{S}$ as a learnable associative memory that continually corrects itself toward the mapping $\bm{k}_t \mapsto \bm{v}_t$.  
The rank-1 update structure, equivalent to a generalized Householder transformation, supports hardware-efficient chunkwise parallelization~\citep{bischof-wy-1987,yang-2024-parallelizing}.  

\paragraph{Gated DeltaNet as Weight Decay.}
Although DeltaNet stabilizes learning, it still retains outdated associations indefinitely.  
Gated DeltaNet (GDN)~\citep{yang-2025-gdn} introduces a scalar forget gate $\brickred{\alpha_t} \in [0,1]$, yielding
\[
\mathbf{S}_t
= \brickred{\alpha_t} (\mathbf{I}-\beta_t\bm{k}_t\bm{k}_t^\top) \mathbf{S}_{t-1}
  + \beta_t\bm{k}_t\bm{v}_t^\top .
\]
Here, $\brickred{\alpha_t}$ acts as a form of \emph{weight decay} on the fast weights \cite{behrouz2025atlas}, implementing a  forgetting mechanism analogous to data-dependent $L_2$ regularization.  
This simple yet effective modification provides a principled way to control memory lifespan and mitigate interference, improving both stability and long-context generalization while preserving DeltaNet’s parallelizable structure.

From this perspective, we observe that GDN can be interpreted as a form of multiplicative positional encoding where the transition matrix is data-dependent and learnable, relaxing the orthogonality constraint of RoPE \citep{yang2025path}.\footnote{When the state transformation matrix preserves its orthogonality, absolute positional encodings can also be applied independently to $\bm{q}$ and $\bm{k}$ to be converted into relative positional encodings during the attention computation \citep{kexuefm-11033}.}

\section{Kimi Delta Attention: Improving Delta Rule with Fine-grained Gating}
We propose Kimi Delta Attention (KDA), a new gated linear attention variant that refines GDN's scalar decay by introducing a fine-grained diagonalized gate $\brickred{\operatorname{Diag}(\boldsymbol{\alpha}_t)}$ that enables fine-grained control over memory decay and positional awareness (as discussed in \S\ref{sec:delta_rule}). We begin by introducing the chunkwise parallelization of KDA, showing how a series of rank-1 matrix transformations can be compressed into a dense representation while maintaining stability under diagonal gating. We then highlight the efficiency gains of KDA over the standard DPLR (\emph{Diagonal-Plus-Low-Rank}) formulation \cite{gu-2022-efficiently,peng-2025-rwkv7}.
\begin{equation}
    \mathbf{S}_t = \left(\mathbf{I}-\beta_t\bm{k}_{t}\bm{k}_{t}^{\top}\right)\brickred{\operatorname{Diag}\left(\bm{\alpha}_t \right)}\mathbf{S}_{t-1} + \beta_t\bm{k}_{t}\bm{v}_{t}^{\top}\in\mathbb{R}^{d_k\times d_v};
    \qquad \bm{o}_t = \mathbf{S}^\top_t \bm{q}_t\in\mathbb{R}^{d_v}
    \label{eq:recurrent_KDA}
\end{equation}
\definecolor{kvcolor}{RGB}{241,140,74}
\begin{figure*}[h!]
    \centering
    \begin{adjustbox}{width=0.9\textwidth}
        \begin{tikzpicture}
            \tikzset{
                state/.style={
                    matrix of nodes,
                    nodes={
                        minimum size=10pt,
                        anchor=center,
                        inner sep=0pt,
                        font=\tiny
                    },
                    nodes in empty cells,
                    inner sep=0pt,
                    rounded corners=1pt
                },
                statelink/.style={
                    dash pattern={on 4pt off 4pt},
                    ->,
                    >={Straight Barb[length=5pt, width=6pt]},
                    line width=1.3pt,
                    shorten >=8pt,
                    shorten <=8pt,
                },
                seplink/.style={
                    dash pattern={on 4pt off 4pt},
                    line width=1pt,
                    shorten >=8pt,
                    shorten <=8pt,
                },
                kvlink/.style={
                    ->,
                    >={Straight Barb[length=5pt, width=6pt]},
                    line width=1.3pt,
                    shorten >=8pt,
                    shorten <=8pt,
                },
            }
            \matrix[state, anchor=west, left delimiter={[}, right delimiter={]}] (st) {
            |[fill=white]| {} & |[fill=midnightblue!20]| {} & |[fill=midnightblue!10]| {} & |[fill=midnightblue!20]| {} \\
            |[fill=midnightblue!60]| {} & |[fill=midnightblue!70]| {} & |[fill=white]| {} & |[fill=midnightblue!40]| {} \\
            |[fill=white]| {} & |[fill=midnightblue!30]| {} & |[fill=midnightblue!50]| {} & |[fill=white]| {} \\
            |[fill=midnightblue!50]| {} & |[fill=midnightblue!20]| {} & |[fill=midnightblue!20]| {} & |[fill=midnightblue!50]| {} \\
            };

            \node[right=10pt of st] (eq) {\large $=$};
            \node[right=0pt of eq.east] (leftbracket) {\large $\Biggl($};
            \matrix[state, right=5pt of leftbracket, left delimiter={[}, right delimiter={]}] (identity) {
            |[fill=black!50]| {} & |[fill=white]| {} & |[fill=white]| {} & |[fill=white]| {} \\
            |[fill=white]| {} & |[fill=black!50]| {} & |[fill=white]| {} & |[fill=white]| {} \\
            |[fill=white]| {} & |[fill=white]| {} & |[fill=black!50]| {} & |[fill=white]| {} \\
            |[fill=white]| {} & |[fill=white]| {} & |[fill=white]| {} & |[fill=black!50]| {} \\
            };
            \node[right=10pt of identity] (minus) {\large $-$};

            \matrix[state,right=10pt of minus] (kvec)
            {
            |[fill=kvcolor!30]| {} \\
            |[fill=kvcolor!50]| {} \\
            |[fill=kvcolor!70]| {} \\
            |[fill=kvcolor!90]| {} \\
            };

            \node[anchor=west, inner sep=0pt] at (kvec.east) {\large $\times$};
            \matrix[state,
            anchor=north west, baseline=0.3cm
            ]
            (kt) at ($(kvec.north east)+(10pt,0)$) {
            |[fill=kvcolor!30]| {} & |[fill=kvcolor!50]| {} & |[fill=kvcolor!70]| {} & |[fill=kvcolor!90]| {} \\
            |[fill=white, opacity=0, minimum size=0.05cm]| {} & |[fill=white, opacity=0, minimum size=0.05cm]| {} & |[fill=white, opacity=0, minimum size=0.05cm]| {} & |[fill=white, opacity=0, minimum size=0.05cm]| {} \\
            };
            \node[right=50pt of kvec] (rightbracket) {\large $\Biggl)$};

            \matrix[state, right=5pt of rightbracket, left delimiter={[}, right delimiter={]}] (decay) {
            |[fill=brickred!80]| {} & |[fill=white]| {} & |[fill=white]| {} & |[fill=white]| {} \\
            |[fill=white]| {} & |[fill=brickred!70]| {} & |[fill=white]| {} & |[fill=white]| {} \\
            |[fill=white]| {} & |[fill=white]| {} & |[fill=brickred!50]| {} & |[fill=white]| {} \\
            |[fill=white]| {} & |[fill=white]| {} & |[fill=white]| {} & |[fill=brickred!60]| {} \\
            };

            \matrix[state, right=20pt of decay, left delimiter={[}, right delimiter={]}] (stm1) {
            |[fill=midnightblue!20]| {} & |[fill=midnightblue!40]| {} & |[fill=midnightblue!60]| {} & |[fill=white]| {} \\
            |[fill=midnightblue!90]| {} & |[fill=midnightblue!70]| {} & |[fill=midnightblue!20]| {} & |[fill=midnightblue!70]| {} \\
            |[fill=midnightblue!30]| {} & |[fill=midnightblue!40]| {} & |[fill=midnightblue!60]| {} & |[fill=midnightblue!50]| {} \\
            |[fill=midnightblue!10]| {} & |[fill=white]| {} & |[fill=midnightblue!80]| {} & |[fill=midnightblue!20]| {} \\
            };

            \node[right=10pt of stm1] (plus) {\large $+$};
            \matrix[state,right=10pt of plus] (kvec)
            {
            |[fill=kvcolor!30]| {} \\
            |[fill=kvcolor!50]| {} \\
            |[fill=kvcolor!70]| {} \\
            |[fill=kvcolor!90]| {} \\
            };

            \node[anchor=west, inner sep=0pt] at (kvec.east) {\large $\times$};
            \matrix[state,
            anchor=north west, baseline=0.3cm
            ]
            (kt) at ($(kvec.north east)+(10pt,0)$) {
            |[fill=kvcolor!30]| {} & |[fill=kvcolor!50]| {} & |[fill=kvcolor!70]| {} & |[fill=kvcolor!90]| {} \\
            |[fill=white, opacity=0, minimum size=0.05cm]| {} & |[fill=white, opacity=0, minimum size=0.05cm]| {} & |[fill=white, opacity=0, minimum size=0.05cm]| {} & |[fill=white, opacity=0, minimum size=0.05cm]| {} \\
            };
            
            \draw[seplink] ($(kvec.north)+(70pt,16pt)$) -- ($(kvec.south)+(70pt,-16pt)$);
            \matrix[state,right=80pt of kvec] (o)
            {
            |[fill=kvcolor!30]| {} \\
            |[fill=kvcolor!50]| {} \\
            |[fill=kvcolor!70]| {} \\
            |[fill=kvcolor!90]| {} \\
            };
            
            \node[right=5pt of o] (eq) {\large $=$};
            \matrix[state, anchor=west, right=10pt of eq, left delimiter={[}, right delimiter={]}] (st) {
            |[fill=white]| {}           & |[fill=midnightblue!60]| {} & |[fill=white]| {}           & |[fill=midnightblue!50]| {} \\
            |[fill=midnightblue!20]| {} & |[fill=midnightblue!70]| {} & |[fill=midnightblue!30]| {} & |[fill=midnightblue!40]| {} \\
            |[fill=midnightblue!10]| {} & |[fill=white]| {}           & |[fill=midnightblue!50]| {} & |[fill=midnightblue!20]| {} \\
            |[fill=midnightblue!20]| {} & |[fill=midnightblue!20]| {} & |[fill=white]| {}           & |[fill=midnightblue!50]| {} \\
            };
            
            \matrix[state,right=10pt of st] (q)
            {
            |[fill=kvcolor!30]| {} \\
            |[fill=kvcolor!50]| {} \\
            |[fill=kvcolor!70]| {} \\
            |[fill=kvcolor!90]| {} \\
            };

        \end{tikzpicture}

    \end{adjustbox}
    \captionsetup{labelformat=empty,labelsep=none}
    \label{fig:KDA-recurrent}
\end{figure*}

\subsection{Hardware-Efficient Chunkwise Algorithm}
\label{sec:kda:chunk}

By partially expanding the recurrence for Eq. \ref{eq:recurrent_KDA} into a chunk-wise formulation, we have:
\begin{equation}
    \begin{aligned}
        \mathbf{S}_{[t]}^r & = \underbrace{\left(\prod_{i=1}^r \left(\mathbf{I} - \beta_{[t]}^i \boldsymbol{k}_{[t]}^i \boldsymbol{k}_{[t]}^{i\top}\right) \brickred{\operatorname{Diag}(\boldsymbol{\alpha}_{[t]}^i)}\right)}_{:= \mathbf{P}_{[t]}^r} \cdot\mathbf{S}_{[t]}^{0} + \underbrace{\sum_{i=1}^{r} \left(\prod_{j=i+1}^r \left(\mathbf{I} - \beta_{[t]}^j \boldsymbol{k}_{[t]}^j \boldsymbol{k}_{[t]}^{j\top}\right)\brickred{\operatorname{Diag}(\boldsymbol{\alpha}_{[t]}^j)}\right)\cdot\beta_{[t]}^i \boldsymbol{k}_{[t]}^i\boldsymbol{v}_{[t]}^{i\top}}_{:=\mathbf{H}_{[t]}^r}
    \end{aligned}
    \label{eq:KDA-recurrent}
\end{equation}

\paragraph{WY Representation}shi
is typically employed to  pack a series rank-1 updates into a single compact representation \citep{bischof-wy-1987}. We follow the formulation of $\mathbf{P}$ in Comba \citep{hu2025comba} to reduce the need for an additional matrix inversion in subsequent computations.
    \begin{align}
        \mathbf{P}_{[t]}^r = \brickred{\operatorname{Diag}(\boldsymbol{\gamma}_{[t]}^r)} - \sum_{i=1}^{r} \brickred{\operatorname{Diag}(\boldsymbol{\gamma}_{[t]}^{i\rightarrow r})} \boldsymbol{k}_{[t]}^i \boldsymbol{w}_{[t]}^{i\top} &&
        \mathbf{H}_{[t]}^r = \sum_{i=1}^{t} \brickred{\operatorname{Diag}\left(\boldsymbol{\gamma}_{[t]}^{i\rightarrow r}\right)} \boldsymbol{k}_{[t]}^i \boldsymbol{u}_{[t]}^{i\top}
        \label{eq:PH_wy}
    \end{align}
    where the auxiliary vector $\boldsymbol{w}_t \in \mathbb{R}^{d_k}$ and $\boldsymbol{u}_t \in \mathbb{R}^{d_v}$ are computed via the following recurrence relation:
    \begin{align}
        \boldsymbol{w}_{[t]}^r &= \beta_{[t]}^r \left( \brickred{\operatorname{Diag}(\boldsymbol{\gamma}_{[t]}^r)} \boldsymbol{k}_{[t]}^r - \sum_{i=1}^{r-1} \boldsymbol{w}_{[t]}^i\left( \boldsymbol{k}_{[t]}^{i\top}\brickred{\operatorname{Diag}\left(\boldsymbol{\gamma}_{[t]}^{i\rightarrow r} \right)}\boldsymbol{k}_{[t]}^r \right)  \right) \\
        \boldsymbol{u}_{[t]}^r &= \beta_{[t]}^r \left(\boldsymbol{v}_{[t]}^r - \sum_{i=1}^{r-1}\boldsymbol{u}_{[t]}^i \left(\boldsymbol{k}_{[t]}^{i\top} \brickred{\operatorname{Diag}\left(\boldsymbol{\gamma}_{[t]}^{i\rightarrow r}\right)} \boldsymbol{k}_{[t]}^r\right)  \right)
    \end{align}

\paragraph{UT transform.}
We apply the UT transform \citep{joffrain-2006-ut} to reduce non-matmul FLOPs, which is crucial to enable better hardware utilization during training.
\begin{align}
\label{eq:gdn-wy}
\mathbf{M}_{[t]}&=\left(\mathbf{I} +  \operatorname{StrictTril} \left(\operatorname{Diag}\left(\beta_{[t]}\right) \left(\brickred{{\bm{\Gamma}}_{[t]}^{1\rightarrow C}} \odot \mathbf{K}_{[t]} \right) \left(\frac{\mathbf{K}_{[t]}}{\brickred{\bm{\Gamma}_{[t]}^{1\rightarrow C}}} \right)^\top\right) \right)^{-1} \operatorname{Diag}\left(\beta_{[t]}\right)\\
\mathbf{W}_{[t]} &= \mathbf{M}_{[t]} \left(\brickred{{\bm{\Gamma}}_{[t]}^{1\rightarrow C}}\odot\mathbf{K}_{[t]}\right),  \quad\quad\quad \mathbf{U}_{[t]}=\mathbf{M}_{[t]} \mathbf{V}_{[t]} 
\end{align}
The inverse of a lower triangular matrix can be efficiently computed through an iterative row-wise approach by forward substitution in Gaussian elimination \citep{grcar_2011}.

Equivalently, in matrix form, we can update the state in chunk-wise:
\begin{equation}
    \mathbf{S}_{[t+1]} = \brickred{\operatorname{Diag}(\boldsymbol{\gamma}_{[t]}^C)} \mathbf{S}_{[t]} +   \left(\brickred{\bm{\Gamma}_{[t]}^{i\rightarrow C}} \odot \mathbf{K}_{[t]}\right)^\top \left(\mathbf{U}_{[t]} - \mathbf{W}_{[t]} \mathbf{S}_{[t]}\right) \in \mathbb{R}^{d_k\times d_v}
\end{equation}
\definecolor{kvcolor}{RGB}{241,140,74}
\definecolor{americanrose}{rgb}{1.0, 0.01, 0.24}
\definecolor{bubblegum}{rgb}{0.99, 0.76, 0.8}
\definecolor{cadmiumred}{rgb}{0.89, 0.0, 0.13}
\begin{figure*}[h]
    \centering
    \begin{adjustbox}{width=0.6\textwidth,center}
    \centering
        \begin{tikzpicture}
            \tikzset{
                state/.style={
                    matrix of nodes,
                    nodes={
                        minimum size=10pt,
                        anchor=center,
                        inner sep=0pt,
                        font=\tiny
                    },
                    nodes in empty cells,
                    inner sep=0pt,
                    rounded corners=1pt
                },
                statelink/.style={
                    dash pattern={on 4pt off 4pt},
                    ->,
                    >={Straight Barb[length=5pt, width=6pt]},
                    line width=1.3pt,
                    shorten >=8pt,
                    shorten <=8pt,
                },
                seplink/.style={
                    dash pattern={on 4pt off 4pt},
                    line width=1pt,
                    shorten >=8pt,
                    shorten <=8pt,
                },
                kvlink/.style={
                    ->,
                    >={Straight Barb[length=5pt, width=6pt]},
                    line width=1.3pt,
                    shorten >=8pt,
                    shorten <=8pt,
                },
            }
            \matrix[state, anchor=west, left delimiter={[}, right delimiter={]}] (stp1) {
            |[fill=white]| {} & |[fill=midnightblue!20]| {} & |[fill=midnightblue!10]| {} & |[fill=midnightblue!20]| {} \\
            |[fill=midnightblue!60]| {} & |[fill=midnightblue!70]| {} & |[fill=white]| {} & |[fill=midnightblue!40]| {} \\
            |[fill=white]| {} & |[fill=midnightblue!30]| {} & |[fill=midnightblue!50]| {} & |[fill=white]| {} \\
            |[fill=midnightblue!50]| {} & |[fill=midnightblue!20]| {} & |[fill=midnightblue!20]| {} & |[fill=midnightblue!50]| {} \\
            };
            \node[right=10pt of stp1] (eq) {\large $=$};
            \matrix[state, right=10pt of eq.east, left delimiter={[}, right delimiter={]}] (decay) {
            |[fill=brickred!80]| {} & |[fill=white]| {} & |[fill=white]| {} & |[fill=white]| {} \\
            |[fill=white]| {} & |[fill=brickred!70]| {} & |[fill=white]| {} & |[fill=white]| {} \\
            |[fill=white]| {} & |[fill=white]| {} & |[fill=brickred!50]| {} & |[fill=white]| {} \\
            |[fill=white]| {} & |[fill=white]| {} & |[fill=white]| {} & |[fill=brickred!60]| {} \\
            };
            \matrix[state, anchor=west, right=20pt of decay, left delimiter={[}, right delimiter={]}] (st) {
            |[fill=midnightblue!20]| {} & |[fill=midnightblue!40]| {} & |[fill=midnightblue!60]| {} & |[fill=white]| {} \\
            |[fill=midnightblue!90]| {} & |[fill=midnightblue!70]| {} & |[fill=midnightblue!20]| {} & |[fill=midnightblue!70]| {} \\
            |[fill=midnightblue!30]| {} & |[fill=midnightblue!40]| {} & |[fill=midnightblue!60]| {} & |[fill=midnightblue!50]| {} \\
            |[fill=midnightblue!10]| {} & |[fill=white]| {} & |[fill=midnightblue!80]| {} & |[fill=midnightblue!20]| {} \\
            };
            \node[right=10pt of st] (plus) {\large $+$};
            \node[right=0pt of plus.east] (leftbracket) {\large $\Biggl($};
            \matrix[state,
            right=3pt of leftbracket
            ]
            (decay1)  {
            |[fill=brickred!80]| {} &|[fill=brickred!50]| {} &|[fill=brickred!40]| {} \\
            |[fill=brickred!70]| {} &|[fill=brickred!70]| {} &|[fill=brickred!80]| {} \\
            |[fill=brickred!50]| {} &|[fill=brickred!90]| {} &|[fill=brickred!60]| {} \\
            |[fill=brickred!60]| {} &|[fill=brickred!80]| {} &|[fill=brickred!90]| {} \\
            };
            \node[anchor=west, inner sep=2pt] (odot) at (decay1.east) {\large $\odot$};
            \matrix[state,
            anchor=west,
            right=0pt of odot
            ]
            (kv)  {
            |[fill=kvcolor!30]| {} &|[fill=kvcolor!30]| {} &|[fill=kvcolor!30]| {} \\
            |[fill=kvcolor!50]| {} &|[fill=kvcolor!50]| {} &|[fill=kvcolor!50]| {} \\
            |[fill=kvcolor!70]| {} &|[fill=kvcolor!70]| {} &|[fill=kvcolor!70]| {} \\
            |[fill=kvcolor!90]| {} &|[fill=kvcolor!90]| {} &|[fill=kvcolor!90]| {} \\
            };
            \node[right=3pt of kv] (rightbracket) {\large $\Biggl)$};
            \matrix[state,
            anchor=west, 
            ]
            (kvt) at ($(kv.east)+(20pt,0)$) {
            |[fill=kvcolor!30]| {} & |[fill=kvcolor!50]| {} & |[fill=kvcolor!70]| {} & |[fill=kvcolor!90]| {} \\
            |[fill=kvcolor!30]| {} & |[fill=kvcolor!50]| {} & |[fill=kvcolor!70]| {} & |[fill=kvcolor!90]| {} \\
            |[fill=kvcolor!30]| {} & |[fill=kvcolor!50]| {} & |[fill=kvcolor!70]| {} & |[fill=kvcolor!90]| {} \\
            };

        \end{tikzpicture}
        \end{adjustbox}

    \captionsetup{labelformat=empty,labelsep=none}
    \label{fig:KDA-chunks}
\end{figure*}
During the output stage, we adopt an inter-block recurrent and intra-block parallel strategy to maximize matrix multiplication throughput, thereby fully utilizing the computational potential of Tensor Cores.
\begin{equation}
\label{eq:gdn-o}
    \mathbf{O}_{[t]} = 
    \underbrace{\left(\brickred{{\bm{\Gamma}}_{[t]}^{1\rightarrow C}} \odot\mathbf{Q}_{[t]}\right)
    \mathbf{S}_{[t]}}_\text{inter chunk} + \underbrace{\operatorname{Tril}\left(\left(\brickred{{\bm{\Gamma}}_{[t]}^{1\rightarrow C}} \odot \mathbf{Q}_{[t]} \right) \left(\frac{\mathbf{K}_{[t]}}{\brickred{{\bm{\Gamma}}_{[t]}^{1\rightarrow C}}} \right)^\top \right)}_\text{intra chunk} \underbrace{\left(\mathbf{U}_{[t]} - \mathbf{W}_{[t]} \mathbf{S}_{[t]}\right)}_{\text{``pseudo''-value term}} \in \mathbb{R}^{C\times d_v}
\end{equation}
\definecolor{kvcolor}{RGB}{241,140,74}
\definecolor{americanrose}{rgb}{1.0, 0.01, 0.24}
\definecolor{bubblegum}{rgb}{0.99, 0.76, 0.8}
\definecolor{cadmiumred}{rgb}{0.89, 0.0, 0.13}
\begin{figure*}[h]
    \centering
    \begin{adjustbox}{width=0.6\textwidth,center}
    \centering
        \begin{tikzpicture}
            \tikzset{
                state/.style={
                    matrix of nodes,
                    nodes={
                        minimum size=10pt,
                        anchor=center,
                        inner sep=0pt,
                        font=\tiny
                    },
                    nodes in empty cells,
                    inner sep=0pt,
                    rounded corners=1pt
                },
                statelink/.style={
                    dash pattern={on 4pt off 4pt},
                    ->,
                    >={Straight Barb[length=5pt, width=6pt]},
                    line width=1.3pt,
                    shorten >=8pt,
                    shorten <=8pt,
                },
                seplink/.style={
                    dash pattern={on 4pt off 4pt},
                    line width=1pt,
                    shorten >=8pt,
                    shorten <=8pt,
                },
                kvlink/.style={
                    ->,
                    >={Straight Barb[length=5pt, width=6pt]},
                    line width=1.3pt,
                    shorten >=8pt,
                    shorten <=8pt,
                },
            }
            
            \matrix[state,
            anchor=north, 
            ]
            (o) at ($(stp1.south)-(0,45pt)$) {
            |[fill=kvcolor!30]| {} & |[fill=kvcolor!50]| {} & |[fill=kvcolor!70]| {} & |[fill=kvcolor!90]| {} \\
            |[fill=kvcolor!30]| {} & |[fill=kvcolor!50]| {} & |[fill=kvcolor!70]| {} & |[fill=kvcolor!90]| {} \\
            |[fill=kvcolor!30]| {} & |[fill=kvcolor!50]| {} & |[fill=kvcolor!70]| {} & |[fill=kvcolor!90]| {} \\
            };
            \node[] at (o-|eq) (eq1) {\large $=$};

            \node[right=0pt of eq1.east] (leftbracket) {\large $\Biggl($};
            \matrix[state,
            right=3pt of leftbracket
            ]
            (decay1)  {
            |[fill=brickred!80]| {} &|[fill=brickred!70]| {} &|[fill=brickred!50]| {} &|[fill=brickred!60]| {} \\
            |[fill=brickred!50]| {} &|[fill=brickred!70]| {} &|[fill=brickred!90]| {} &|[fill=brickred!80]| {} \\
            |[fill=brickred!40]| {} &|[fill=brickred!80]| {} &|[fill=brickred!60]| {} &|[fill=brickred!90]| {} \\
            };
            \node[anchor=west, inner sep=2pt] (odot) at (decay1.east) {\large $\odot$};
            \matrix[state,
            anchor=west,
            right=0pt of odot
            ]
            (q)  {
            |[fill=kvcolor!30]| {} &|[fill=kvcolor!50]| {} &|[fill=kvcolor!70]| {} &|[fill=kvcolor!90]| {} \\
            |[fill=kvcolor!30]| {} &|[fill=kvcolor!50]| {} &|[fill=kvcolor!70]| {} &|[fill=kvcolor!90]| {} \\
            |[fill=kvcolor!30]| {} &|[fill=kvcolor!50]| {} &|[fill=kvcolor!70]| {} &|[fill=kvcolor!90]| {} \\
            };
            \node[right=3pt of q] (rightbracket) {\large $\Biggl)$};
            \matrix[state, right=5pt of rightbracket, left delimiter={[}, right delimiter={]}] (stp1) {
            |[fill=white]| {} & |[fill=midnightblue!20]| {} & |[fill=midnightblue!10]| {} & |[fill=midnightblue!20]| {} \\
            |[fill=midnightblue!60]| {} & |[fill=midnightblue!70]| {} & |[fill=white]| {} & |[fill=midnightblue!40]| {} \\
            |[fill=white]| {} & |[fill=midnightblue!30]| {} & |[fill=midnightblue!50]| {} & |[fill=white]| {} \\
            |[fill=midnightblue!50]| {} & |[fill=midnightblue!20]| {} & |[fill=midnightblue!20]| {} & |[fill=midnightblue!50]| {} \\
            };
            \node[right=10pt of stp1] (plus) {\large $+$};
            
            \matrix[state, right=10pt of plus, left delimiter={[}, right delimiter={]}] (attn) {
            |[fill=cadmiumred!70]| {} & |[fill=white]| {} & |[fill=white]| {} \\
            |[fill=cadmiumred!40]| {} & |[fill=cadmiumred!80]| {} & |[fill=white]| {} \\
            |[fill=cadmiumred!50]| {} & |[fill=cadmiumred!30]| {} & |[fill=cadmiumred!60]| {} \\
            };
            \matrix[state,
            anchor=west, 
            ]
            (kvt) at ($(attn.east)+(10pt,0)$) {
            |[fill=kvcolor!30]| {} & |[fill=kvcolor!50]| {} & |[fill=kvcolor!70]| {} & |[fill=kvcolor!90]| {} \\
            |[fill=kvcolor!30]| {} & |[fill=kvcolor!50]| {} & |[fill=kvcolor!70]| {} & |[fill=kvcolor!90]| {} \\
            |[fill=kvcolor!30]| {} & |[fill=kvcolor!50]| {} & |[fill=kvcolor!70]| {} & |[fill=kvcolor!90]| {} \\
            };

        \end{tikzpicture}
        \end{adjustbox}

    \captionsetup{labelformat=empty,labelsep=none}
    \label{fig:KDA-chunko}
\end{figure*}

\subsection{Efficiency Analysis}
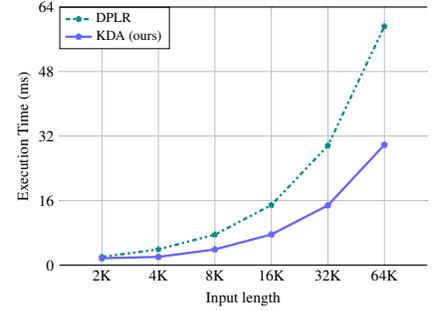
\begin{wrapfigure}[14]{r}{0.34\textwidth}
    \centering
    \vspace{-2em}
    \resizebox{\linewidth}{!}{
        \begin{tikzpicture}
            \begin{axis}[
                    trim axis left,
                    trim axis right,
                    ymajorgrids=true,
                    xmajorgrids=true,
                    tickwidth=0pt,
                    tick align=inside,
                    xlabel=Input length,
                    enlarge x limits=0.15,
                    width=10cm, height=7.5cm,
                    ymin=0, ymax=64,
                    symbolic x coords={
                            2K,4K,8K,16K,32K,64K
                        },
                    ytick={0,16,32,48,64},
                    yticklabels={0,16,32,48,64},
                    xlabel near ticks,
                    ylabel=Execution Time (ms),
                    ylabel style={at={(0.05,0.5)}},
                    axis line style={opacity=0},
                    legend style={
                            at={(0,1)},
                            anchor=north west,
                            legend cell align=left,
                            font=\small,
                        },
                ]

                \addplot[
                    line width=1.5pt,
                    dashdotdotted,
                    mark=star,
                    mark size=2pt,
                    mark options={scale=1},
                    color=darkcyan
                ] plot coordinates {
                        (2K,  2.021408 )
                        (4K,  3.900816 )
                        (8K,  7.550000 )
                        (16K, 14.894096)
                        (32K, 29.636353)
                        (64K, 59.221313)
                    };
                \addlegendentry{DPLR}

                \addplot[
                    line width=1.5pt,
                    mark=pentagon*,
                    draw=blue!60,,
                    mark options={
                            fill=blue!60,
                            fill opacity=1.0,
                            solid
                        },
                    mark size=1.5pt,
                    opacity=1.0,
                ] plot coordinates {
                        (2K,    1.698688)
                        (4K,    2.026816)
                        (8K,    3.877184)
                        (16K,   7.599360)
                        (32K,  14.811120)
                        (64K,  29.831713)
                    };
                \addlegendentry{\text{KDA (ours)}}

                \draw [semithick, black] (axis description cs:0,0) -- (axis description cs:1,0); 
                \draw [semithick, black] (axis description cs:0,1) -- (axis description cs:1,1); 

            \end{axis}

        \end{tikzpicture}
    }

    \caption{
        Execution time of kernels for varying input lengths, with a uniform batch size of 1 and 16 heads.
    }
    \label{fig:kernel}
\end{wrapfigure}
In terms of representational capacity, KDA aligns with the generalized DPLR formulation, i.e., $\mathbf{S}_t = (\mathbf{D} - \bm{a}_t \bm{b}_t^{\top}) \mathbf{S}_{t-1} + \bm{k}_t \bm{v}_t^{\top}$, both exhibiting fine-grained decay behavior. However, such fine-grained decay introduces numerical precision issues during division operations (e.g., the intra-chunk computation in Eq.~\ref{eq:gdn-o}). To address this, prior work such as GLA~\citep{yang-etal-2024-gla} performs computations in the logarithmic domain and introduces secondary chunking in full precision. This approach, however, prevents full utilization of half-precision matrix multiplications and significantly reduces operator speed.
By binding both variables $\bm a$ and $\bm b$ to $\bm k$, KDA effectively alleviates this bottleneck—reducing the number of second-level chunk matrix computations from four to two, and further eliminating three additional matrix multiplications. As a result, the operator efficiency of KDA improves by roughly 100\% compared to the DPLR formulation. A detailed analysis is provided in \S \ref{sec:related_dplr}.

\input{figures/model}

\section{The Kimi Linear Model Architecture}
The main backbone of our model architecture follows Moonlight \citep{liu-2025-moonlight}.
In addition to fine-grained gating, we also leverage several components to further improve the expressiveness of Kimi Linear.
The overall Kimi Linear architecture is shown in Figure \ref{fig:scaling-model}.

\paragraph{Neural Parameterization}
Let $\bm{x}_t \in \mathbb{R}^d$ be the $t$-th token input representation, the input to KDA for each
head $h$ is computed as follows
\begin{align*}
\bm{q}^h_t,\bm{k}^h_t &= \operatorname{L2Norm}(\operatorname{Swish}(\operatorname{ShortConv}(\mathbf{W}^h_{q/k}\bm{x}_t)))\in \mathbb{R}^{d_k}\\
\bm{v}^h_t &= \operatorname{Swish}(\operatorname{ShortConv}(\mathbf{W}^h_v\bm{x}_t))\in \mathbb{R}^{d_v} \\
\brickred{\bm{\alpha}^h_t} &= f(\mathbf{W}_{\alpha}^{\uparrow}\mathbf{W}_{\alpha}^{\downarrow}\bm{x}_t) \in [0,1]^{d_k}\\
\beta^h_t &= \operatorname{Sigmoid}(\mathbf{W}_{\beta}^h\bm{x}_t) \in [0,1]\\
\end{align*}
where $d_k, d_v$ represent the key and value head dimensions, which are set to 128 for all experiments.
For $\bm{q},\bm{k},\bm{v}$, we apply a $\operatorname{ShortConv}$ followed by a $\operatorname{Swish}$ activation, following \citep{yang-2025-gdn}. 
The $\bm{q}$ and $\bm{k}$ representations are further normalized using $\operatorname{L2Norm}$ to ensure eigenvalues stability, as suggested by \cite{yang-2024-parallelizing}. 
The per-channel decay $\brickred{\bm{\alpha}^h_t}$ is parameterized via a low-rank projection (\(\mathbf{W}_{\alpha}^{\downarrow}\) and \(\mathbf{W}_{\alpha}^{\uparrow}\) with rank equal to the head dimension) and a decay function $f(\cdot)$ similar to those used in GDN and Mamba~\citep{yang-2025-gdn,mamba2}. 
Before the output projection through $\mathbf{W}_o \in \mathbb{R}^{d \times d}$, we use a head-wise RMSNorm \citep{zhang2019root} and a data-dependent gating mechanism \citep{qiu2025gated} parameterized as:
\begin{equation}
\begin{aligned}
\bm{o}_t = \mathbf{W}_o\left( \operatorname{Sigmoid}\left(\mathbf{W}_g^{\uparrow}\mathbf{W}_g^{\downarrow} \bm{x}_t\right)\odot \operatorname{RMSNorm}\left(\operatorname{KDA}\left( \bm{q}_t,\bm{k}_t,\bm{v}_t,\brickred{\bm{\alpha}_t},\beta_t \right) \right)\right) 
\end{aligned}
\end{equation}
Here, the output gate adopts a low-rank parameterization similar to the forget gate, to ensure a fair parameter comparison, while maintaining performance comparable to full-rank gating and alleviating the Attention Sink \citep{qiu2025gated}. 
The choice of nonlinear activation function is further discussed in \S\ref{sec:ablation}.

\paragraph{Hybrid model architecture}
Long‑context retrieval remains the primary bottleneck for pure linear attention, we therefore hybridize KDA with a small number of full global‑attention (Full MLA) layers \cite{deepseekaiv3}. 
For Kimi Linear, we chose a layerwise approach (alternating entire layers) over a headwise one (mixing heads within layers) for its superior infrastructure simplicity and training stability. 
Empirically, a uniform 3:1 ratio, i.e., repeating 3 KDA layers to 1 full MLA layer, provided the best quality–throughput trade‑off.
We discuss other hybridization strategies in \S~\ref{sec:hybrid}.

\paragraph{No Position Encoding (NoPE) for MLA Layers.}
In Kimi Linear, we apply NoPE to all full attention (MLA) layers.
This design delegates the entire responsibility for encoding positional information and recency bias (see \S~\ref{sec:delta_rule}) to the KDA layers.
KDA is thus established as the primary position-aware operator, fulfilling a role analogous to, or arguably stronger than, auxiliary components like short convolutions \citep{allen2025physics} or SWA \citep{puvvada2025swangpt}.
Our findings align with prior results \citep{yang2025ropenopeagainnew,barbero2025round,deepseekaiv3}, who similarly demonstrated that complementing global NoPE attention with a dedicated position-aware mechanism yields competitive long-context performance.

We note that NoPE offers practical advantages, particularly for MLA. 
First, NoPE enables their conversion to the highly-efficient pure Multi-Query Attention (MQA) during inference. 
Second, it simplifies long-context training, as it obviates the need for RoPE parameter adjustments, such as frequency base tuning or methods like YaRN \citep{peng2023yarn}.

\section{Experiments}

\subsection{Synthetic tests}
We start by evaluating KDA against other competing linear attention methods on three synthetic tasks, serving as benchmark tests for long-context performance.
Across all experiments, we adopt a consistent model configuration of 2 layers with 2 attention heads, each having a head dimension of 128.
For each task, we train the model for at most 20,000 steps with a grid search over learning rates in $\{5\times 10^{-5}, 1\times 10^{-4}, 5\times 10^{-4}, 1\times 10^{-3}\}$.
We then present the best-performing training accuracy curves. Specifically, we compare two scenarios: (1) the performance of different tasks as training length increases from 256 to 2,048 tokens, measuring the peak accuracy; and (2) the convergence speed of KDA, GDN, and Mamba2 with a fixed context length of 1,024 tokens.

\input{figures/syn}
\paragraph{Palindrome} Palindrome requires the model to reproduce a given sequence of random tokens in reverse order. 
As illustrated in Table~\ref{tab:palindrome}, given an input like ``\text{O G R S U N E}'',  the model must generate its exact reversal.
Such copying tasks are known to be difficult for linear attention models \citep{jelassi-2024-repeat}, as they struggle to precisely retrieve the entire history from a compressed, fixed-size memory state.
\begin{center}
    \begin{tabular}{rcccccccccccccccccc}
        \textbf{Input}  & O      & G      & R      & S      & U      & N      & E      & \raisebox{0.5pt}{\textcolor{gray}{\texttt{\textsc{<sep>}}}} & E & N & U & S & R & G & O      \\
        \textbf{Output} & $\phi$ & $\phi$ & $\phi$ & $\phi$ & $\phi$ & $\phi$ & $\phi$ & $\phi$                                                      & N & U & S & R & G & O & $\phi$ \\
    \end{tabular}
    \label{tab:palindrome}
\end{center}

\paragraph{Multi Query Associative Recall (MQAR)} MQAR assesses the model's ability to retrieve values associated with multiple queries that appear at various positions within the context.
For instance, as shown in Table~\ref{tab:mqar}, the model is asked to recall \brickred{0} for the query \brickred{B} and \midnightblue{5} for \midnightblue{G}.
This task is known to be highly correlated with language modeling performance \citep{arora-2023-zoology}.

\begin{center}
    \begin{tabular}{rcccccccccccccccccc}
        \textbf{Input}  & A      & 1      & C      & 3      & \brickred{B} & \brickred{0} & M      & 8      & \midnightblue{G} & \midnightblue{5} & E      & 4      & \raisebox{0.5pt}{\textcolor{gray}{\texttt{\textsc{<sep>}}}} & \brickred{B} & \midnightblue{G} \\
        \textbf{Output} & $\phi$ & $\phi$ & $\phi$ & $\phi$ & $\phi$       & $\phi$       & $\phi$ & $\phi$ & $\phi$           & $\phi$           & $\phi$ & $\phi$ & $\phi$                                                      & \brickred{0} & \midnightblue{5} \\
    \end{tabular}
    \label{tab:mqar}
\end{center}

\paragraph{Stack} We assess the state tracking capabilities \citep{grazzi2025unlocking} of each candidate by simulating the standard LIFO (Last In First Out) stack operations.
Our setup involves 64 independent stacks, each identified by a unique ID.
The model processes a sequence of two operations: 1) PUSH: an action like ``\raisebox{0.5pt}{\textcolor{gray}{\texttt{\textsc{<push>}}}} \midnightblue{1} \midnightblue{G}'' adds the element \midnightblue{G} to stack \midnightblue{1}; 2) POP: an action like ``\raisebox{0.5pt}{\textcolor{gray}{\texttt{\textsc{<pop>}}}} \brickred{0} \brickred{E}'' requires the model to predict the element \brickred{E} most recently pushed onto stack \brickred{0}. 
The objective is to accurately track the states of all stacks and predict the correct element upon each pop request.

Figure~\ref{fig:syn} shows the final results. 
Across all tasks, KDA consistently achieves the highest accuracy as the sequence length increases from 256 to 2,048 tokens.
In particular, on the Palindrome and recall-intensive MQAR tasks, KDA converges significantly faster than GDN. 
This confirms the benefits of our fine-grained decay, which enables the model to selectively forget irrelevant information while preserving crucial memories more precisely.
We also observe that Mamba2~\citep{mamba2}, a typical linear attention that uses only multiplicative decay and lacks a delta rule, fails on all tasks in our model settings.

\subsection{Ablation on Key Components of Kimi Linear}
\label{sec:ablation}

\begin{table}[t]
    \centering
    \small
    \renewcommand{\arraystretch}{1.1}
    \caption{
        Ablation study on the hybrid ratio of KDA to MLA attention and other key components.
        We list the training and validation perplexities (lower is better) for comparison.
        The best-performing model, used in our final experiments, is highlighted in gray.
    }
    \setlength{\tabcolsep}{8pt}
    \begin{tabular}{rrccc}
        \toprule
                                                                     &      & Training PPL ($\downarrow$) & Validation PPL ($\downarrow$) \\
        \midrule
        \rowcolor{gray!20} \multirow{5}{*}{Hybrid ratio}             & 3:1  & \textbf{9.23}               & \textbf{5.65}                 \\
                                                                     & 0:1  & 9.45                        & 5.77                          \\
                                                                     & 1:1  & 9.29                        & 5.66                          \\
                                                                     & 7:1  & 9.23                        & 5.70                          \\
                                                                     & 15:1 & 9.34                        & 5.82                          \\
        \midrule
        \multicolumn{2}{r}{\textit{w/o} output gate}                 & 9.25 & 5.67                                                        \\
        \multicolumn{2}{r}{\textit{w/ $\mathrm{swish}$} output gate} & 9.43 & 5.81                                                        \\
        \multicolumn{2}{r}{\textit{w/o} convolution layer}           & 9.29 & 5.70                                                        \\
        \bottomrule
    \end{tabular}
    \label{tab:ablation}
\end{table}

We conducted a series of ablation studies by directly comparing different models to the first-scale scaling law model, i.e., 16 heads, 16 layers. 
All models were trained with the same FLOPs budget and hyperparameters for a fair comparison.
We report the training and validation perplexities (PPLs) in Table~\ref{tab:ablation}.
The validation PPL is calculated on a high-quality dataset whose distribution differs significantly from the pre-training corpus, emphasizing generalization under distribution shift, and thus the differences in training and validation perplexities. 

\paragraph{Output gate}
We compare our default $\mathrm{Sigmoid}$ output gate against two variants: one with no gating and another with $\mathrm{swish}$ gating. 
The results show that removing the gate degrades performance. 
Moreover, the $\mathrm{swish}$ gate adopted by \cite{yang-2025-gdn} performs substantially worse than $\mathrm{Sigmoid}$. Our observation is consistent with \cite{qiu2025gated}, who also conclude that $\mathrm{Sigmoid}$ gating offers superior performance.
So we adopt $\mathrm{Sigmoid}$ across all of our experiments, including GDN-H.

\paragraph{Convolution Layer} 
Lightweight depthwise convolutions with a small kernel size (e.g., 4) can be effective at capturing local token dependencies \citep{allen2025physics} and are widely adopted by many recent architectures \citep{mamba2,arora-2023-zoology,yang-2024-parallelizing}. 
We validate its efficacy in Table~\ref{tab:ablation}, demonstrating that convolutional layers continue to play a non-negligible role in hybrid models.

\paragraph{Hybrid ratio}
We performed an ablation study to determine the optimal hybrid ratio of KDA linear attention layers to MLA full attention layers. Among the configurations tested, the 3:1 ratio (3 KDA layers for every 1 MLA layer) yielded the best results, achieving the lowest training and validation losses.
We observed clear trade-offs with other ratios: a higher ratio (e.g., 7:1) produced a comparable training loss but led to significantly worse validation performance, while a lower ratio (e.g., 1:1) maintained a similar validation loss but at the cost of increased inference overhead. 
Furthermore, the pure full-attention baseline (0:1) performed poorly. Thus, the 3:1 configuration offers the most effective balance between model performance and computational efficiency.

\paragraph{NoPE vs. RoPE}

As shown in Table~\ref{tab:long_ctx}, the Kimi Linear consistently excels on long‑context evaluations, whereas Kimi Linear (RoPE) attains similar scores on short‑context tasks. We posit that this divergence arises from how positional bias is distributed across depth. In Kimi Linear (RoPE), the global attention layer carries a strong, explicit relative positional signal, while the linear attention (e.g., GDN) contributes a weaker, implicit positional inductive bias. This mismatch yields an overemphasis on short‑range order in the global layer, which benefits short contexts but makes the model less flexible when adapting mid‑training to extended contexts.
By contrast, Kimi Linear induces a more balanced positional bias across layers, which improves robustness and extrapolation at long ranges, leading to stronger long‑context performance. 
Regarding long context performance, as shown in Table~\ref{tab:long_ctx}, Kimi Linear achieves the best average score across different long context benchmarks, which verifies the benefits we claim in the last section.

\subsection{Scaling Law of Kimi Linear}
\pgfdeclareplotmark{redstar}{
    \node[star,star point ratio=2.25,minimum size=6pt,
          inner sep=0pt,draw=black,solid,fill=brickred!80] {};
}
\pgfdeclareplotmark{bluestar}{
    \node[star,star point ratio=2.25,minimum size=6pt,
          inner sep=0pt,draw=black,solid,fill=midnightblue!80] {};
}

\begin{table}[t!]
    \centering
    \small
    \renewcommand{\arraystretch}{1.1}
    \begin{threeparttable}
    \caption{
    Model configurations and hyperparameters for scaling law experiments. 
    }
    \setlength{\tabcolsep}{7pt}
    \label{tab:scaling_params}
    \begin{tabular}{cccccccc}
        \toprule
        \# Act. Params.\tnote{$\dagger$} & Head & Layer & Hidden & Tokens & lr & batch size\tnote{$\ddagger$} \\
        \midrule
        653M  & 16 & 16 & 1216 & \textcolor{white}{0}38.8B & $2.006\times 10^{-3}$ & 336 \\
        878M  & 18 & 18 & 1376 & \textcolor{white}{0}59.8B & $1.790\times 10^{-3}$ & 432 \\
        1.1B  & 20 & 20 & 1536 & \textcolor{white}{0}85.2B & $1.617\times 10^{-3}$ & 512 \\
        1.4B  & 22 & 22 & 1632 & 102.5B                    & $1.486\times 10^{-3}$ & 576 \\
        1.7B  & 24 & 24 & 1776 & 128.0B                    & $1.371\times 10^{-3}$ & 640 \\
        \bottomrule
    \end{tabular}
    \begin{tablenotes}
        \small
        \item[$\dagger$] Denotes the number of activated parameters in our MoE models, excluding embeddings.
        \item[$\ddagger$] All models were trained with a context length of 4,096.
    \end{tablenotes}
\end{threeparttable}
\end{table}

\begin{figure}[t!]
    \centering
    \small 
    \resizebox{0.5\textwidth}{!}{
    \begin{tikzpicture}
        \begin{axis}[
            xmode=log,
            ymode=log,
            log ticks with fixed point,
            x filter/.code=\pgfmathparse{#1},
            xlabel={PFLOP/s-days},
            ylabel style={at={(0.5,0.0)}},
            ylabel={Loss},
            ylabel style={at={(0.05,0.5)}},
            xmin=1.3,
            xmax=18,
            ymin=1.96,
            ymax=2.27,
            xtick={1, 10},
            xticklabels={{$10^0$}, {$10^1$}}, 
            ytick={2.0, 2.1, 2.2},
            yticklabels={$2.0$, $2.1$, $2.2$}, 
            width=9cm, height=8cm,
            scaled y ticks=false,
            grid=both,
            legend style={ at={(1,1)}, anchor=north east, legend cell align=left, font=\scriptsize, },
        ]
            \addplot[midnightblue!80, dashed, line width=1.2pt, domain=1.3:18]  {
            2.30919 * exp(-0.05364 * ln(x))
            };
            \addlegendentry{MLA: $2.3092 \times C^{-0.0536}$}

            \addplot[brickred!80, dashed, line width=1.2pt, domain=1.3:18] {
            2.28789 * exp(-0.05265 * ln(x))
            };
            \addlegendentry{Kimi Linear: $2.2879 \times C^{-0.0527}$}

            \addplot[color=brickred!80, only marks, mark=redstar]
            coordinates { 
            (1.7561361621634257, 2.2256522972637383)
            (3.6389031312011673, 2.130743119369714)
            (6.761923198438021, 2.066681874261965)
            (9.873512038478644, 2.030533820622505)
            (15.437117088796946, 1.9826816360711195)
            };

            \addplot[color=midnightblue!80, only marks, mark=bluestar] coordinates
            { 
            (1.697379395880878, 2.249736827606759)
            (3.498749644306514, 2.1514321567048675)
            (6.4511332473629, 2.0874647828148913)
            (9.386512491492146, 2.0503538819031504)
            (14.583615044168363, 2.002056205240299)
            };
            \draw [->, thick, black] (axis cs:5.0841580520520635, 2.1) --  (axis cs:5.880916850890163, 2.1)
                node[
                    midway,   
                    xshift=30pt,
                    above=5pt,
                    sloped,     
                    font=\small,
                    black       
                ] 
                {\large$1.16 \times$};
        \end{axis}
    \end{tikzpicture}
    }
    \caption{The fitted scaling law curves for MLA and Kimi Linear. 
    }
    \label{fig:scaling_law}
\end{figure}
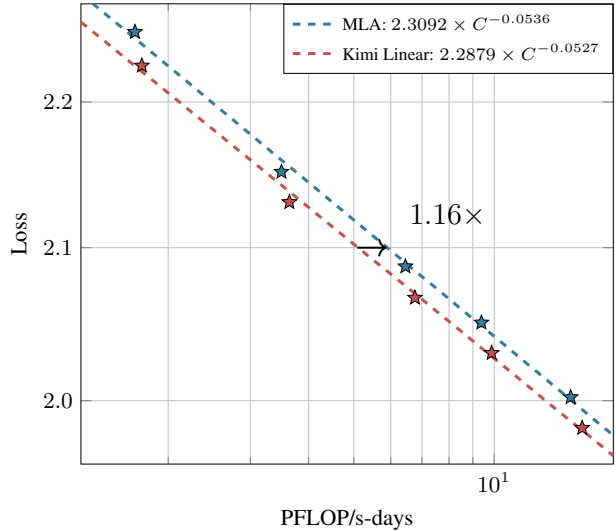

We conducted scaling law experiments on a series of MoE models following the Moonlight \citep{liu-2025-moonlight} architecture. In all experiments, we activated 8 out of 64 experts and utilized the Muon optimizer \citep{liu-2025-moonlight}.
Details and hyperparameters are listed in Table~\ref{tab:scaling_params}.

For MLA, following the Chinchilla scaling law methodology \citep{hoffmann-2022-chinchilla}, we trained five language models of different sizes, carefully tuning their hyperparameters through grid search to ensure optimal performance for each model.
For KDA, we maintained the best hybrid ratio of 3:1 as ablated in Table~\ref{tab:ablation}.
Except for this, we adhered strictly to the MLA training configuration without any modifications.
As shown in Figure~\ref{fig:scaling_law}, Kimi Linear achieves $\sim 1.16\times$ computational efficiency compared to the MLA baselines with compute optimal training. We expect that careful hyperparameter tuning will yield superior scaling curves for KDA.

\subsection{Experimental Setup}

\paragraph{Kimi Linear and baselines settings} 

We evaluate our Kimi Linear model against a full-attention MLA baseline and a hybrid Gated DeltaNet (GDN-H) baseline, all of which share the same architecture, parameter count, and training setup for fair comparisons. The model configuration is largely aligned with Moonlight \citep{liu-2025-moonlight}, with the key distinction that MoE sparsity is increased to 32. Each model activates 8 out of 256 experts, including one shared expert,  resulting in 48 billion total parameters and 3 billion active parameters per forward pass. 
The first layer is implemented as a dense layer without MoE, ensuring stable training. To evaluate the effectiveness of NoPE in Kimi Linear, we also introduce a hybrid KDA baseline using RoPE with the same model configuration, referred to as Kimi Linear (RoPE).

\paragraph{Evaluation Benchmarks} Our evaluation encompasses three primary categories of benchmarks, each designed to assess distinct capabilities of the model:

\begin{itemize}[leftmargin=12pt]
    \item \textbf{Language Understanding and Reasoning}: Hellaswag \citep{zellers-2019-hellaswag}, ARC-Challenge \citep{clark-2018-arc}, Winogrande \citep{keisuke-2019-winogrande}, MMLU \citep{hendrycks-2021-mmlu}, TriviaQA \citep{joshi2017triviaqa}, MMLU-Redux \citep{gema2024we}, MMLU-Pro \citep{wang2024mmlu}, GPQA-Diamond \citep{rein2024gpqa}, BBH \citep{suzgun2022challenging}, and 
    \citep{white2024livebench}.
    \item \textbf{Code Generation}: LiveCodeBench v6 \footnote{Questions from 2024.8 to 2025.5}\citep{jain2024livecodebench}, EvalPlus \citep{evalplus}.
    \item  \textbf{Math \& Reasoning}: AIME 2025, MATH 500, HMMT 2025, PolyMath-en.
    \item  \textbf{Long-context}: MRCR \footnote{\url{https://huggingface.co/datasets/openai/mrcr}} , RULER \citep{hsieh2024ruler}, Frames \citep{krishna2024fact}, HELMET-ICL \citep{yen2025helmet}, RepoQA \citep{liu2024repoqa}, Long Code Arena \citep{bogomolov2024long} and LongBench v2 \citep{bai2024longbench}.
    \item \textbf{Chinese Language Understanding and Reasoning}:  C-Eval \citep{huang2023c}, and CMMLU \citep{li-etal-2024-cmmlu}.
\end{itemize}

\paragraph{Evaluation Configurations} All models are evaluated using temperature 1.0. 
For benchmarks with high variance, we report the score of Avg@$k$. For base model, We employ perplexity-based evaluation for MMLU, MMLU-Redux, GPQA-Diamond, and C-Eval. Otherwise, generation-based evaluation is adopted. To mitigate the high variance inherent to GPQA-Diamond, we report the mean score across eight independent runs. All evaluations are conducted using our internal framework derived from LM-Harness-Evaluation \citep{biderman2024lessons}, ensuring consistent settings across all models.

\subsubsection{Pre-training recipe}
\paragraph{Pre-training recipe} 
All models are pretrained using a 4,096-token context window, the MuonClip optimizer, and the WSD learning rate schedule, processing a shared total of 1.4 trillion tokens sampled from the K2 pretraining corpus \citep{kimi2025k2}. The learning rate is set to $1.1\times 10^{-3}$, and the global batch size is fixed at 32 million tokens. They also adopt the same annealing schedule and long-context activation phase established in Kimi K2 \citep{kimi2025k2}.

Our final released Kimi Linear checkpoint is pretrained using the same procedure, but with an expanded total of 5.7 trillion tokens to match the pretraining tokens of Moonlight. In addition, the final checkpoint supports a context length of up to 1 million tokens. We compare the performance of Kimi Linear\string@5.7T and Moonlight in Appendix~\ref{appendix:results}

\subsubsection{Post-training recipe}

\paragraph{SFT recipe}

The SFT dataset extends the Kimi K2 \citep{kimi2025k2} SFT data by incorporating additional reasoning tasks, creating a large-scale instruction-tuning dataset that spans diverse domains with a heavy emphasis on math and coding. We employ a multi-stage SFT approach, initially training the model on a broad range of diverse SFT data for general instruction-following, followed by scheduled targeted training on reasoning-intensive data to enhance the model’s reasoning capabilities.

\paragraph{RL recipe}
For the RL training prompt set, we primarily integrate three data sources: mathematics, code, and STEM. The main purpose of this enhancement is to boost the model’s reasoning ability. Before conducting RL, we pre-selected data that matches a moderate difficulty level for the starting checkpoint.

A known risk of RL training is the potential degeneration of general capabilities. To mitigate this, we incorporate the PTX loss~\cite{ouyang2022training} during RL, following the practice of K2~\cite{kimi2025k2}. This involves concurrent SFT on a high-quality, distributionally diverse dataset in the RL progress. Our PTX dataset spans both reasoning and general-purpose tasks. All data mentioned above are subsets derived from the training recipe of the K2 model~\cite{kimi2025k2}.

For the RL algorithm, we use the same algorithm as in  K1.5~\cite{kimiteam2025kimik15scalingreinforcement}, while introducing several advanced tricks. We noticed that the precision mismatch between training and inference engines may lead to unstable RL learning. Therefore we introduce  truncated importance sampling, a method that effectively mitigates the policy mismatch between rollout and training~\cite{yao2025offpolicy}. We also dynamically adjust the KL penalty and the mini batch size (\textit{i.e.}, the number of updates per iteration) to make the RL training stable and avoid collapse of entropy~\cite{cui2025entropy}.

\subsection{Main results} 

\subsubsection{Kimi Linear\string@1.4T results}

\paragraph{Pretrain results}

We compared our Kimi Linear model against two baselines (MLA and hybrid GDN-H) using a 1.4T pretraining corpus in Table~\ref{tab:pretrain}. 
The evaluation focused on three areas: general knowledge, reasoning (math and code), and Chinese tasks.
Kimi Linear consistently outperformed both baselines across almost all categories.
\begin{itemize}[leftmargin=12pt]
    \item General Knowledge: Kimi Linear scores highest on all of the key benchmarks like BBH, MMLU and HellaSwag.
    \item Reasoning: It leads in math (GSM8K) and most code tasks (CRUXEval). However, it scores slightly lower on EvalPlus compared to GDN-H.
    \item Chinese Tasks: Kimi Linear achieves the top scores on CEval and CMMLU.
\end{itemize}
In summary, Kimi Linear demonstrated the strongest performance, positioning it as a strong alternative to full-attention architectures at short context pretraining.

\begin{table}[!ht]
    \centering
    \small
    \renewcommand{\arraystretch}{1.1}
    \caption{Performance comparison of Kimi Linear with the full-attention MLA baseline and the hybrid GDN baseline, all after the same pretraining recipe. Kimi Linear consistently outperforms both MLA and GDN-H on short-context pretrain evaluations. Best per-column results are \textbf{bolded}.}
    \setlength{\tabcolsep}{12pt}
    \begin{tabular}{@{}r l c c c c}
        \toprule
                                               &    Type Base            & MLA           & GDN-H         & Kimi Linear   \\
        \midrule
                                               & Trained Tokens & 1.4T          & 1.4T          & 1.4T          \\
        \midrule
        \multirow{7}{*}{\textit{General}}
                                               & HellaSwag      & 81.7          & 82.2          & \textbf{82.9} \\
                                               & ARC-challenge  & 64.6          & 66.5          & \textbf{67.3} \\
                                               & Winogrande     & 78.1          & 77.9          & \textbf{78.6} \\
                                               & BBH            & 71.6          & 70.6          & \textbf{72.9} \\
                                               & MMLU           & 71.6          & 72.2          & \textbf{73.8} \\
                                               & MMLU-Pro       & 47.2          & 47.9          & \textbf{51.0} \\
                                               & TriviaQA       & 68.9          & 70.1          & \textbf{71.7} \\
        \midrule

        \multirow{5}{*}{\textit{Math \& Code}} & GSM8K          & 83.7          & 81.7          & \textbf{83.9} \\
                                               & MATH           & \textbf{54.7} & 54.1          & \textbf{54.7} \\
                                               & EvalPlus       & 59.5          & \textbf{63.1} & 60.2          \\
                                               & CRUXEval-I-cot & 51.6          & 56.0          & \textbf{56.6} \\
                                               & CRUXEval-O-cot & 61.5          & 58.1          & \textbf{62.0} \\

        \midrule
        \multirow{2}{*}{\textit{Chinese}}      & CEval          & 79.3          & 79.1          & \textbf{79.5} \\
                                               & CMMLU          & 79.5          & 80.7          & \textbf{80.8} \\

        \bottomrule
    \end{tabular}

    \label{tab:pretrain}
\end{table}

\begin{table}[h]
    \centering
    \footnotesize
    \setlength{\tabcolsep}{9pt}
    \caption{Performance comparison of Kimi Linear with the full-attention MLA baseline and the hybrid GDN baseline, all using the same SFT recipe after pretraining. Kimi Linear consistently outperforms both MLA and GDN-H on short-context instruction-tuned benchmarks. Best per-column results are \textbf{bolded}.}
    \label{tab:instruct-1t-eval}
    \begin{tabular}{@{}r l c c c}
        \toprule
         &      Type Instruct                     & MLA           & GDN-H         & Kimi Linear   \\
        \midrule
         & Trained Tokens            & 1.4T          & 1.4T          & 1.4T          \\
        \midrule
        \multirow{6}{*}{\textit{General}}
         & BBH                       & 68.2          & 68.5          & \textbf{69.4} \\
         & MMLU                      & 75.7          & 75.6          & \textbf{77.0} \\
         & MMLU-Pro                  & 65.7          & 64.8          & \textbf{67.4} \\
         & MMLU-Redux                & 79.2          & 78.7          & \textbf{80.3} \\
         & GPQA-Diamond (Avg@8)      & 57.1          & 58.6          & \textbf{62.1} \\
         & LiveBench (Pass@1)        & 45.7          & \textbf{46.4} & 45.2          \\
        \midrule
        \multirow{6}{*}{\textit{Math \& Code}}
         & AIME 2025 (Avg@64)        & 20.6          & 21.1          & \textbf{21.3} \\
         & MATH500 (Acc.)            & 80.8          & \textbf{83.0} & 81.2          \\
         & HMMT 2025 (Avg@32)        & 11.3          & 11.3          & \textbf{12.5} \\
         & PolyMath-en (Avg@4)       & 41.3          & 41.5          & \textbf{43.6} \\
         & LiveCodeBench v6 (Pass@1) & 25.1          & 25.4          & \textbf{26.0} \\
         & EvalPlus                  & \textbf{62.6} & 62.5          & 61.0          \\

        \bottomrule
    \end{tabular}
\end{table}

\begin{table}[t]
    \centering
    \small
    \renewcommand{\arraystretch}{1.1}
    \caption{
        Comparisons of Kimi Linear with MLA, GDN-H, and Kimi Linear (RoPE) across long-context benchmarks.
        The last column reports the overall average ($\uparrow$). All models is trained on 1.4T tokens. Best per-column results are \textbf{bolded}.
    }
    \label{tab:long_ctx}
    \begin{adjustbox}{width=1\columnwidth, center}
        \setlength{\tabcolsep}{7pt}
        \begin{tabular}{lccccccccccccccc|c}
            \toprule
                               & \multirow{2}{*}{RULER} & \multirow{2}{*}{MRCR} & \multirow{2}{*}{HELMET-ICL} & \multirow{2}{*}{LongBench V2} & \multirow{2}{*}{Frames} & \multirow{2}{*}{RepoQA} & \multicolumn{2}{c}{Long Code Arena} & \multirow{2}{*}{Avg.}                 \\
            \cmidrule{8-9}
                               &                        &                       &                             &                               &                         &                         & Lib                                 & Commit                &               \\
            \midrule
            MLA                & 81.3                   & 22.6                  & 88.0                        & \textbf{36.1}                 & \textbf{60.5}           & 63.0                    & 32.8                                & \textbf{33.2}         & 52.2          \\
            GDN-H              & 80.5                   & 23.9                  & 85.5                        & 32.6                          & 58.7                    & 63.0                    & 34.7                                & 30.5                  & 51.2          \\
            Kimi Linear (RoPE) & 78.8                   & 22.0                  & 88.0                        & 35.4                          & 59.9                    & 66.5                    & 31.3                                & 32.5                  & 51.8          \\
            \rowcolor{gray!15}
            Kimi Linear        & \textbf{84.3}          & \textbf{29.6}         & \textbf{90.0}               & 35.0                          & 58.8                    & \textbf{68.5}           & \textbf{37.1}                       & 32.7                  & \textbf{54.5} \\
            \bottomrule
        \end{tabular}
    \end{adjustbox}
\end{table}

\paragraph{SFT results}

Kimi Linear demonstrates strong performance across both general and math \& code tasks after undergoing the same supervised fine-tuning (SFT) recipe, consistently outperforming MLA and GDN-H. 
In general tasks, Kimi Linear leads across the board, achieving the top scores on various MMLU benchmarks, BBH, and GPQA-Diamond.
In math \& code tasks, it surpasses both baselines on difficult benchmarks like AIME 2025, HMMT 2025, PolyMath-en, and LiveCodeBench.
Despite some minor exceptions like MATH500 and EvalPlus, Kimi Linear shows robust superiority across the tasks, confirming its clear superiority to the other models tested (GDN-H and MLA).

\paragraph{Long Context Performance Evaluation}
We evaluate the long-context performance of Kimi Linear against three baseline models—MLA, GDN-H, and Kimi Linear (RoPE)—across several benchmarks at 128k context length (see Table~\ref{tab:long_ctx}). The results highlight Kimi Linear's clear superiority in these long-context tasks.
It consistently outperformed MLA and GDN-H, achieving the highest scores on RULER (84.3) and RepoQA (68.5) by a significant margin. This pattern of outperformance held across most other tasks, except for LongBench V2 and Frames. 
Overall, Kimi Linear achieved the highest average score (54.5), further reinforcing its effectiveness as a leading attention architecture in long-context scenarios.

\paragraph{RL results}
\input{figures/rl_KDA_mla}

To compare the RL convergence properties of Kimi Linear and MLA, we conduct RLVR using the in-house mathematics training set from \citep{kimi2025k2}, and evaluate on mathematics test sets (e.g., AIME 2025, MATH500), while keeping the algorithm and all hyperparameters identical to ensure a fair comparison of performance.

As shown in Figure \ref{fig:rl_compare}, Kimi Linear demonstrates better efficiency compared to MLA. On the training set, even though both models start at similar points, the growth rate of training accuracy for Kimi Linear is significantly higher than that of MLA, and the gap gradually widens. On the test set, similar phenomena are observed. For example, on MATH500 and AIME2025, Kimi Linear achieves faster and better improvement compared to MLA. Overall, in reasoning-intensive long-form generation under RL, we empirically observe that Kimi Linear performs significantly better than MLA.

\textbf{Summary of overall findings} During the pretraining and SFT stages, a clear performance hierarchy was established: Kimi Linear outperformed GDN-H, which in turn outperformed MLA. 
However, this hierarchy shifted in long-context evaluations. While Kimi Linear maintained its top position, GDN-H's performance declined, placing it behind MLA. 
Furthermore, in the RL stage, Kimi Linear also demonstrated superior performance over MLA. Overall, Kimi Linear consistently ranked as the top performer across all stages, establishing itself as a superior alternative to full attention architectures.

\subsection{Efficiency Comparison}
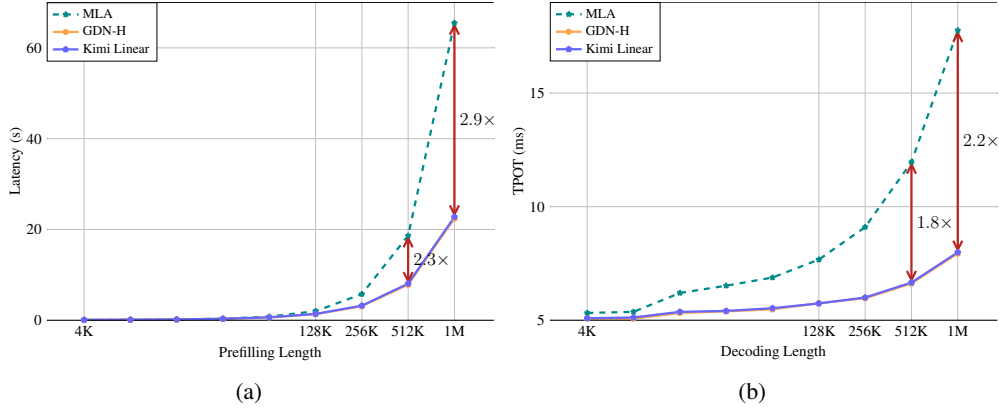
\begin{figure}[h!]
    \centering
    \begin{subfigure}[b]{0.4\textwidth}
        \centering
        \resizebox{\textwidth}{!}{
            \begin{tikzpicture}
                \begin{axis}[
                        trim axis left,
                        trim axis right,
                        xmode=log,
                        log basis x={10},
                        ymajorgrids=true,
                        xmajorgrids=true,
                        tickwidth=0pt,
                        tick align=inside,
                        xlabel={Prefilling Length},
                        enlarge x limits=0.1,
                        width=12cm, height=9cm,
                        xmin=0,xmax=1080000,
                        ymin=0, ymax=70,
                        scaled x ticks=false,
                        scaled y ticks=false,
                        xtick={4096,131072,262144,524288,1048576},
                        xticklabels={4K,128K,256K,512K,1M},
                        ytick={0,20,40,60},
                        yticklabels={0,20,40,60},
                        xlabel near ticks,
                        ylabel=Latency (s),
                        ylabel style={at={(0.05,0.5)}},
                        axis line style={opacity=0},
                        legend style={
                                at={(0,1)},
                                anchor=north west,
                                legend cell align=left,
                                font=\small,
                            },
                    ]

                    \addplot[
                        line width=1.5pt,
                        dash pattern={on 4pt off 4pt},
                        mark=star,
                        mark size=2pt,
                        mark options={scale=1},
                        color=darkcyan
                    ] plot coordinates {
                        (4096, 0.064)
                        (8192, 0.099)
                        (16384, 0.182)
                        (32768, 0.359)
                        (65536, 0.796)
                        (131072, 2.005)
                        (262144, 5.764)
                        (524288, 18.629)
                        (1048576, 65.460)
                        };
                    \addlegendentry{MLA}
                    \addplot[
                        line width=1.5pt,
                        mark=10-pointed star,
                        mark size=2pt,
                        draw=orange!70,
                    ] plot coordinates {
                        (4096, 0.099)
                        (8192, 0.123)
                        (16384, 0.184)
                        (32768, 0.324)
                        (65536, 0.624)
                        (131072, 1.341)
                        (262144, 3.101)
                        (524288, 7.847)
                        (1048576, 22.415)
                        };
                    \addlegendentry{GDN-H}

                    \addplot[
                        line width=1.5pt,
                        mark=pentagon*,
                        draw=blue!60, 
                        mark options={
                                fill=blue!60, 
                                fill opacity=1.0,
                                solid
                            },
                        mark size=1.5pt,
                        opacity=1.0,
                    ] plot coordinates {
                        (4096, 0.096)
                        (8192, 0.125)
                        (16384, 0.186)
                        (32768, 0.327)
                        (65536, 0.643)
                        (131072, 1.397)
                        (262144, 3.214)
                        (524288, 8.068)
                        (1048576, 22.753)
                        };
                    \addlegendentry{\text{Kimi Linear}}

                    \draw [semithick, black] (axis description cs:0,0) -- (axis description cs:1,0); 
                    \draw [semithick, black] (axis description cs:0,1) -- (axis description cs:1,1); 
                    
                    \draw[<->,>={Straight Barb[length=6pt, width=6pt]}, line width=1.5pt, brickred] (axis cs:1048576,22.753) -- (axis cs:1048576,65.460);
                    \draw[<->,>={Straight Barb[length=6pt, width=6pt]}, line width=1.5pt, brickred] (axis cs:524288,8.068) -- (axis cs:524288,18.629);
                    \node[right, black] at (axis cs:1048576, 44.1065) (1M) {\large${2.9\times}$};
                    \node[right, black] at (axis cs:524288, 13.3485) (512K) {\large${2.3\times}$};
                \end{axis}

            \end{tikzpicture}
        }
        \caption{}
        \label{fig:prefilling}
    \end{subfigure}
    \begin{subfigure}[b]{0.4\textwidth}
        \centering
        \resizebox{\textwidth}{!}{
            \begin{tikzpicture}
                \begin{axis}[
                        trim axis left,
                        trim axis right,
                        xmode=log,
                        log basis x={10},
                        ymajorgrids=true,
                        xmajorgrids=true,
                        tickwidth=0pt,
                        tick align=inside,
                        xlabel={Decoding Length},
                        enlarge x limits=0.1,
                        width=12cm, height=9cm,
                        xmin=0,xmax=1080000,
                        ymin=5, ymax=19,
                        scaled x ticks=false,
                        scaled y ticks=false,
                        xtick={4096,131072,262144,524288,1048576},
                        xticklabels={4K,128K,256K,512K,1M},
                        ytick={5,10,15},
                        yticklabels={5,10,15},
                        xlabel near ticks,
                        ylabel=TPOT (ms),
                        ylabel style={at={(0.05,0.5)}},
                        axis line style={opacity=0},
                        legend style={
                                at={(0,1)},
                                anchor=north west,
                                legend cell align=left,
                                font=\small,
                            },
                    ]

                    \addplot[
                        line width=1.5pt,
                        dash pattern={on 4pt off 4pt},
                        mark=star,
                        mark size=2pt,
                        mark options={scale=1},
                        color=darkcyan
                    ] plot coordinates {
                        (4096, 5.32)
                        (8192, 5.37)
                        (16384, 6.2)
                        (32768, 6.52)
                        (65536, 6.88)
                        (131072, 7.67)
                        (262144, 9.1)
                        (524288, 11.96)
                        (1048576, 17.76)
                        };
                    \addlegendentry{MLA}

                    \addplot[
                        line width=1.5pt,
                        mark=10-pointed star,
                        mark size=2pt,
                        draw=orange!70,
                    ] plot coordinates {
                        (4096, 5.07)
                        (8192, 5.07)
                        (16384, 5.32)
                        (32768, 5.39)
                        (65536, 5.48)
                        (131072, 5.75)
                        (262144, 5.97)
                        (524288, 6.62)
                        (1048576, 7.94)
                        };
                    \addlegendentry{GDN-H}
                    \addplot[
                        line width=1.5pt,
                        mark=pentagon*,
                        draw=blue!60, 
                        mark options={
                                fill=blue!60, 
                                fill opacity=1.0,
                                solid
                            },
                        mark size=1.5pt,
                        opacity=1.0,
                    ] plot coordinates {
                        (4096, 5.09)
                        (8192, 5.12)
                        (16384, 5.37)
                        (32768, 5.41)
                        (65536, 5.53)
                        (131072, 5.74)
                        (262144, 6.0)
                        (524288, 6.66)
                        (1048576, 7.99)
                        };
                    \addlegendentry{\text{Kimi Linear}}
                    
                    \draw [semithick, black] (axis description cs:0,0) -- (axis description cs:1,0); 
                    \draw [semithick, black] (axis description cs:0,1) -- (axis description cs:1,1); 
                    
                    \draw[<->,>={Straight Barb[length=6pt, width=6pt]}, line width=1.5pt, brickred] (axis cs:1048576,7.99) -- (axis cs:1048576,17.76);
                    \node[right, black] at (axis cs:1048576, 12.9) (1M) {\large${2.2\times}$};
                    
                    \draw[<->,>={Straight Barb[length=6pt, width=6pt]}, line width=1.5pt, brickred] (axis cs:524288,6.66) -- (axis cs:524288,11.96);
                    \node[right, black] at (axis cs:524288, 9.3) (512K) {\large${1.8\times}$};
                \end{axis}

            \end{tikzpicture}
        }
        \caption{}
        \label{fig:decoding}
    \end{subfigure}

    \caption{
        (\subref{fig:prefilling}) The prefilling time of MLA (full attention), hybrid GDN-H and our Kimi Linear.
        (\subref{fig:decoding}) The time per output token (TPOT) for MLA, GDN-H and Kimi Linear during decoding. (We use batch size = 1 here for tests.)
    }
\end{figure} 

\paragraph{Prefilling \& Decoding speed}
We compare the training and decoding times for full attention MLA \citep{deepseekaiv3}, GDN-H, and Kimi Linear in Figure~\ref{fig:prefilling} and Figure~\ref{fig:decoding}. Note that all models are based on the Kimi Linear 48B setting, with the same number of layers and attention heads. 
We observe that:
1) Despite incorporating a more fine-grained decay mechanism, Kimi Linear introduces negligible latency overhead compared to GDN-H during prefilling.
As shown in Figure \ref{fig:prefilling}, their performance curves are virtually indistinguishable, confirming that our method maintains high efficiency.
The hybrid Kimi Linear model demonstrates a clear efficiency advantage over the MLA baseline as sequence length increases. 
While its performance is comparable to MLA at shorter lengths (4k–16k), it becomes significantly faster from 128k onwards. 
This efficiency gap widens dramatically at scale, with Kimi Linear outperforming MLA by a factor of $2.3$ for 512k sequences and $2.9$ for 1M sequences.
As shown in Figure~\ref{fig:decoding-best},  Kimi Linear fully demonstrates its advantages during the decoding phase. For decoding at 1M context length, Kimi Linear is $6\times$ faster than full attention.

\section{Discussions}

\subsection{Kimi Delta Attention as learnable position embeddings}\label{sec:delta_rule}

The standard attention in transformers is by design agnostic to the sequence order of its inputs \citep{vaswani-2017-attention}, thus necessitating explicit positional encodings \citep{press-2022-alibi, shaw2018selfattention}. 
Among various methods, RoPE \citep{su2024roformer} has emerged as the \textit{de facto} standard in modern LLMs due to its effectiveness \citep{touvron-2023-llama,agarwal2025gpt,deepseekaiv3}.
The mechanism of multiplicative positional encodings like RoPE can be analyzed through a generalized attention formulation:
\begin{equation}
    s_{t,i} = \bm{q}_t^{\top}  \left( \prod_{j=i+1}^t \mathbf{R}_j\right) \bm{k}_i \label{eq:posemb}
\end{equation} 

where the position relationship between the $t$-th query $\bm{q}_t$ and the $i$-th key $\bm{k}_i$ is reflected by the cumulative matrix products. RoPE defines the transformation matrix $\mathbf{R}_j$ as a block diagonal matrix composed of $d_k/2$ 2D rotation matrices $\mathbf{R}_j^{k} = \begin{psmallmatrix} \cos(j\theta_k) & -\sin (j\theta_k) \\ \sin(j\theta_k) & \cos(j\theta_k) \end{psmallmatrix}$ with \textbf{per-2-dimensional} angular frequency $\theta_k$. Due to the properties of rotation matrices, i.e., $\mathbf{R}_{t-i}=\mathbf{R}_t^\top\mathbf{R}_i$, absolute positional information $\mathbf{R}_t$ and $\mathbf{R}_i$ can be applied separately to $\bm{q}_t$ and $\bm{k}_i$, which are then transformed into relative positional information $t-i$ encoded as $\prod_{j=i+1}^t \mathbf{R}_j=\begin{psmallmatrix} \cos((t-i)\theta_k) & -\sin ((t-i)\theta_k) \\ \sin((t-i)\theta_k) & \cos((t-i)\theta_k) \end{psmallmatrix}$.

Consequently, we show that linear attentions with the gated delta rule can be expressed in a comparable formulation in Eq.~\ref{eq:gdnpos}.
Similar forms for other attention variants are summarized in Table~\ref{tab:KDA-parallel}.
\begin{align}\label{eq:gdnpos}
    \bm{o}_t = \sum_{i=1}^t \left( \bm{q}_t^{\top} \left(\prod_{j=i+1}^t\brickred{\mathbf{A}_j}\left(\mathbf{I}-\beta_j\bm{k}_j\bm{k}_j^\top\right) \right)\bm{k}_j\right) \bm{v}_j 
\end{align}

\begin{table}[t!]
    \centering
    \small
    \caption{
        An overview of attention mechanisms in their mathematically equivalent recurrent ($\boldsymbol{o}_t$) and parallel ($\mathbf{O}$) forms.
        We omitted the normalization term and $\beta_t$ to achieve a more concise representation.
        The function \(\phi\) refers to the infinite-dimensional feature space corresponding to the exponential kernel, i.e., \(\phi(\bm{q})^\top \phi(\bm{k}) = \exp(\bm{q}^\top \bm{k})\).
    }
    \renewcommand{\arraystretch}{1.4}
    \begin{adjustbox}{width=1\columnwidth}
        \setlength{\tabcolsep}{4pt}
        \begin{tabular}{rlll}
            \toprule
                                                                                                                                                                                                                              \rowcolor{white} & Recurrent form                                                                                                                                                 & Parallel form \\
            \midrule
            SA \citep{vaswani-2017-attention}                                                                                                                                                                                  &
            $\sum\limits_{j=1}^t  \exp\left( \bm{q}_t^{\top} \bm{k}_j \right)\bm{v}_j $                                                                                                                                        &
            $\left(\exp\left( \mathbf{QK^\top} \right)\odot \mathbf{M}\right) \mathbf{V}$                                                                                                                                                                                                                                                                                                                       \\
            SA + RoPE \citep{su2024roformer}                                                                                                                                                                                   &
            $\sum\limits_{j=1}^t \exp\left( \bm{q}_t^{\top} {\left( \prod\limits_{s=j+1}^t \bm{R}_s\right)} \bm{k}_j \right) \bm{v}_j $                                                                                        &
            $\left(\exp\left( \mathbf{R\left(Q\right)R\left(K\right)^\top} \right)\odot \mathbf{M}\right) \mathbf{V}$                                                                                                                                                                                                                                                                                           \\
            LA \citep{wang-2020-linformer}                                                                                                                                                                                     &
            $\sum\limits_{j=1}^t \left( \bm{q}_t^{\top} \bm{k}_j \right)\bm{v}_j  $                                                                                                                                            &
            $\left(\mathbf{QK^\top} \odot \mathbf{M}\right) \mathbf{V}$                                                                                                                                                                                                                                                                                                                                         \\
            Mamba2 \citep{mamba2}                                                                                                                                                                                              &
            $\sum\limits_{j=1}^t \left( \bm{q}_t^{\top} {\left(\prod\limits_{s=j+1}^t\brickred{\alpha_s} \right)}\bm{k}_j \right)\bm{v}_j  $                                                                                   &
            $\left( \mathbf{QK^\top} \odot \brickred{\mathcal{A}}\odot \mathbf{M}\right)\mathbf{V}$                                                                                                                                                                                                                                                                                                             \\
            GLA \citep{yang-etal-2024-gla}                                                                                                                                                                                     &
            $\sum\limits_{j=1}^t \left( \bm{q}_t^{\top} {\left(\prod\limits_{s=j+1}^t\brickred{\operatorname{Diag}\left(\bm{\alpha}_s\right)} \right)}\bm{k}_j \right)\bm{v}_j  $                                              & $\left(\left( \mathbf{Q} \odot \brickred{\mathbf{\Gamma}}\right) \left(\frac{\mathbf{K}}{\brickred{\mathbf{\Gamma}}} \right)^\top\odot \mathbf{M}\right)\mathbf{V}$                 \\
            DeltaNet \citep{schlag-2021-deltanet}                                                                                                                                                                           &
            $\sum\limits_{j=1}^t \left( \bm{q}_t^{\top} {\left(\prod\limits_{s=j+1}^t\left(\mathbf{I}-\bm{k}_s\bm{k}_s^\top\right) \right)}\bm{k}_j \right)\bm{v}_j  $                                                         &
            $\left(\mathbf{QK^\top} \odot \mathbf{M}\right)\left( \mathbf{I} + \mathbf{KK^{\top}}\odot \mathbf{M}^{-} \right)^{-1}\mathbf{V}$                                                                                                                                                                                                                                                                   \\
            FoX \citep{lin2025forgetting}                                                                                                                                                                                      &
            $\sum\limits_{j=1}^t \exp\left( \bm{q}_t^{\top} \bm{k}_j \right){\left(\prod\limits_{s=j+1}^t \brickred{\alpha_s} \right)}\bm{v}_j  $                                                                              &
            $\left(\exp\left( \mathbf{QK^\top} \right)\odot \brickred{\mathcal{A}} \odot \mathbf{M} \right) \mathbf{V}$                                                                                                                                                                                                                                                                                         \\
            DeltaFormer \citep{zhong2025understanding}                                                                                                                                                                         &
            $\sum\limits_{j=1}^t  \left(\phi(\bm{q}_t)^{\top} {\left(\prod\limits_{s=j+1}^t\left(\mathbf{I}-\phi\left(\bm{k}_s\right)\phi\left(\bm{w}_s\right)^\top\right) \right)}\phi\left(\bm{k}_j\right)\right) \bm{v}_j
            $                                    &
            $\left(\exp\left( \mathbf{QK^\top} \right)\odot \mathbf{M}\right) \left( \mathbf{I} + \exp\left(\mathbf{WK^{\top}} \right)\odot \mathbf{M}^-\right)^{-1}\mathbf{V}$
            \\
            PaTH-FoX \citep{yang2025path}                                                                                                                                                                                          &
            $\sum\limits_{j=1}^t \exp\left( \bm{q}_t^{\top} {\left(\prod\limits_{s=j+1}^t\left(\mathbf{I}-\bm{w}_s\bm{w}_s^\top\right) \right)}\bm{k}_j \right)\left(\prod\limits_{s=j+1}^t\brickred{\alpha_s}\right) \bm{v}_j $                                  &
            $\left(\exp\left(\left(\mathbf{QK^\top} \odot \mathbf{M}\right) \left(\mathbf{I} + \mathbf{WW^{\top}} \odot \mathbf{M}^-\right)^{-1}\right)\odot \brickred{\mathcal{A}}\odot\mathbf{M}\right)\mathbf{V}$                                                                                                                                                                                            \\

            GDN \citep{yang-2025-gdn}                                                                                                                                                                                          &
            $\sum\limits_{j=1}^t \left( \bm{q}_t^{\top} {\left(\prod\limits_{s=j+1}^t\brickred{\alpha_s}\left(\mathbf{I}-\bm{k}_s\bm{k}_s^\top\right) \right)}\bm{k}_j \right)\bm{v}_j  $                                      &
            $\left(\mathbf{QK^\top} \odot \brickred{\mathcal{A}}\odot \mathbf{M}\right)\left( \mathbf{I} + \mathbf{KK^{\top}}\odot \brickred{\mathcal{A}} \odot \mathbf{M}^{-}\right)^{-1}\mathbf{V}$                                                                                                                                                                                                           \\
            Comba \citep{hu2025comba}                                                                                                                                                                                          &
            $\sum\limits_{j=1}^t \left( \bm{q}_t^{\top} {\left(\prod\limits_{s=j+1}^t\left(\brickred{\alpha_s}-\bm{k}_s\bm{k}_s^\top\right) \right)}\bm{k}_j \right)\bm{v}_j  $                                                &
            $\left(\mathbf{QK^\top} \odot \brickred{\mathcal{A}}\odot \mathbf{M}\right)\left( \mathbf{I} + \mathbf{KK^{\top}}\odot \brickred{\mathcal{A}^{i-1/j}} \odot \mathbf{M}^{-}\right)^{-1}\mathbf{V}$                                                                                                                                                                                                   \\
            RWKV7 \citep{peng-2025-rwkv7}                                                                                                                                                                                      &
            $\sum\limits_{j=1}^t \left( \bm{q}_t^{\top} {\left(\prod\limits_{s=j+1}^t\left(\brickred{\operatorname{Diag}\left(\bm{\alpha}_s\right)}-\left(\bm{b}_s\odot\hat{\bm{k}}_s\right)\hat{\bm{k}}_s^\top\right) \right)}\bm{k}_j \right)\bm{v}_j  $           &
            $\left(\left(\mathbf{Q} \odot \brickred{\mathbf{\Gamma}}\right)\left(\frac{{\mathbf{K}}}{\brickred{\mathbf{\Gamma}}} \right)^\top\odot \mathbf{M}\right)\left( \mathbf{I} + \left(\hat{\mathbf{K}} \odot \brickred{\overset{0 \to t-1}{\mathbf{\Gamma}}}\right)\left(\frac{\tilde{\mathbf{K}}\odot \mathbf{B}}{\brickred{\mathbf{\Gamma}}} \right)^\top\odot\mathbf{M}^{-1}\right)^{-1}\mathbf{V}$                                  \\

            \rowcolor{gray!20} \textbf{KDA} (ours)                                                                                                                                                                             &
            $\sum\limits_{j=1}^t \left( \bm{q}_t^{\top} {\left(\prod\limits_{s=j+1}^t\brickred{\operatorname{Diag}\left(\bm{\alpha}_s\right)}\left(\mathbf{I}-\bm{k}_s\bm{k}_s^\top\right) \right)}\bm{k}_j \right) \bm{v}_j $ &
            $\left(\left(\mathbf{Q} \odot \brickred{\mathbf{\Gamma}}\right)\left(\frac{\mathbf{K}}{\brickred{\mathbf{\Gamma}}} \right)^\top\odot \mathbf{M}\right)\left( \mathbf{I} + \left(\mathbf{K} \odot \brickred{\mathbf{\Gamma}}\right)\left(\frac{\mathbf{K}}{\brickred{\mathbf{\Gamma}}} \right)^\top\odot\mathbf{M}^{-1}\right)^{-1}\mathbf{V}$\\
            \bottomrule
        \end{tabular}
    \end{adjustbox}
    \label{tab:KDA-parallel}
\end{table}

From this perspective, GDN can be interpreted as a form of multiplicative positional encoding whose transition matrix is data-dependent, thereby relaxing the orthogonality constraint imposed by RoPE and can be potentially more powerful \citep{yang2025path}.
\footnote{
When preserving orthogonality, absolute positional encodings can be applied independently to $\bm{q}$ and $\bm{k}$, which are then automatically transformed into relative positional encodings during the attention computation \citep{kexuefm-11033}.
}
This provides a potential solution to the known extrapolation issues of RoPE, whose fixed frequencies can cause overfitting to context lengths seen during training \citep{xiong-2023-llamalong,peng2023yarn}.
Some recent works adopt workarounds like partial RoPE \citep{barbero2025round} or even forgo explicit positional encodings entirely (NoPE) \citep{kazemnejad2023impact,puvvada2025swangpt, deepseekaiv3}. 
Given that GDN serves as an analogue role to RoPE, we choose NoPE for global full attention layers (MLA) in our model, allowing positional information to be captured dynamically by our proposed KDA model. 

Moreover, a key strength of RoPE is its fine-grained positional encoding, achieved by assigning different rotation frequencies to each pair of dimensions, which functions analogously to a Nonuniform Fourier Transform \citep{barbero2025round,hua2024fourier} along the feature dimension.
Standard GDN, however, employs a per-head scalar decay and lacks this per-dimensional diversity, which motivates us to propose KDA with a learnable channel-wise gate.

\newenvironment{longlisting}{\captionsetup{type=listing,labelfont=bf}}{}
\renewcommand{\theFancyVerbLine}{\ttfamily\textcolor[rgb]{0.5,0.5,0.5}{\scriptsize\arabic{FancyVerbLine}}}

\setminted{
    fontsize=\footnotesize,
    fontfamily=tt,
    linenos,
    frame=lines,
    breaklines,
    numbersep=1.5pt,
}
\begin{figure}[h]
\begin{subfigure}[t]{0.48\textwidth}
\begin{minted}[
    fontsize=\scriptsize,
    highlightlines={6,13-16,25-27,31-32},
    highlightcolor=red!5,
    ]{python}
def chunk_dplr(q, k, v, a, b, g, chunk_size):
    B, H, T, K, V, BT = *q.shape, v.shape[-1], chunk_size
    NT, S = T // BT, k.new_zeros(B, H, K, V)
    q, k, v, a, b, g = map(lambda x: rearrange(x, 'b h (n c) d -> b h n c d', c=BT), [q, k, v, a, b, g])
    gc = g.cumsum(-2)
-   Aab, Aak, Aqb, Aqk = (torch.zeros(B, H, NT, BT, BT) for _ in range(4))

    for i in range(BT):
        a_i, q_i, g_i = (x[:,:,:,i,None] for x in (a, q, gc))
        mask = (torch.arange(BT) <= i)[..., None]
        s1_i = (g_i - gc).exp().where(mask, 0)
        s2_i= (g_i - g[:,:,:,i,None] - gc).where(mask, 0)
-       Aqk[..., i, :] = (q_i * k * s1_i).sum(-1)
-       Aqb[..., i, :] = (q_i * b * s1_i).sum(-1)
-       Aab[..., i, :] = (a_i * b * s2_i).sum(-1)
-       Aak[..., i, :] = (a_i * k * s2_i).sum(-1)
    for i in range(1, BT):
        Aab[..., i, :i] = Aab[..., i, :i] + (Aab[..., i, :, None] * Aab[..., :, :i]).sum(-2)
    Aab = Aab + torch.eye(BT)
    u, w = Aab @ (Aak @ v), Aab @ ((gc-g).exp() * a)
    o = torch.zeros_like(v)
    mask = torch.triu(torch.ones(BT, BT), diagonal=1)
    for i in range(0, NT):
        q_i, k_i, v_i, u_i, w_i, b_i = (x[:, :, i] for x in (q, k, v, u, w, b))
-       o1 = Aqk[:, :, i] @ v_i
-       o2 = Aqb[:, :, i] @ (u_i + w_i @ S)
-       o3 = (q_i * gkc[:, :, i].exp()) @ S
        o[:, :, i] = o1 + o2 + o3
        decay = (gc[:, :, i, -1, None] - gc[:, :, i]).exp()
        S = S * gc[:,:,i,-1,:,None].exp()
-       S +=(k_i * decay).transpose(-1,-2) @ v_i
-       S +=(b_i * decay).transpose(-1,-2) @ (u_i + w_i @ S)
    return o, S
\end{minted}    
\caption{PyTorch-style pseudo code for chunkwise DPLR. }
\label{listing:dplr}
\end{subfigure}
\hfill
\begin{subfigure}[t]{0.48\textwidth}
\begin{minted}[
    fontsize=\scriptsize,
    highlightlines={6,14-15,26,29},
    highlightcolor=green!10,
    ]{python}
def chunk_kda(q, k, v, a, b, g, chunk_size):
    B, H, T, K, V, BT = *q.shape, v.shape[-1], chunk_size
    NT, S = T // BT, k.new_zeros(B, H, K, V)
    q, k, v, g = map(lambda x: rearrange(x, 'b h (n c) ... -> b h n c ...', c=BT), [q, k, v, g])
    gc = g.cumsum(-2)
+   Aqk, Akk = (torch.zeros(B, H, NT, BT, BT) for _ in range(2))

    for i in range(BT):
        k_i, q_i = k[:, :, :, i, None], q[:, :, :, i, None]
        g_i = gc[...,i:i+1,:]
        mask = (torch.arange(BT) <= i)[..., None]
        s1_i = (g_i - gc).exp().where(mask, 0)
        s2_i = (gc - g_i).exp()
+       Aqk[:, :, :, i, :] = (q_i * k * s1_i).sum(-1)
+       Akk[..., i] = (k_i * k * s2_i).sum(-1)
    mask = torch.triu(torch.ones(BT, BT), diagonal=0)
    A = -Akk.masked_fill(mask, 0)
    for i in range(1, BT):
        A[..., i, :i] = A[..., i, :i] + (A[..., i, :, None] * A[..., :, :i].clone()).sum(-2)
    A = (A + torch.eye(BT))
    w, u = A @ (gc.exp() * k), A @ v
    o = torch.zeros_like(v)
    mask = torch.triu(torch.ones(BT, BT), diagonal=1)
    for i in range(0, NT):
        q_i, k_i, u_i, g_i, w_i = (x[:, :, i] for x in (q, k, u, gc, w))
+       o[:,:,i]=(q_i *g_i.exp()) @ S + Aqk @(u_i-w_i @ S)
        decay = (g_i[:,:,-1:] - g_i).exp()
        S = S * g_i[:, :, -1, :, None].exp()
+       S += (k_i * decay).transpose(-1,-2) @ v_i
    return o, S
\end{minted}   
\caption{PyTorch-style pseudo code for chunkwise KDA. }
\label{listing:KDA} 
\end{subfigure}
\label{listing:KDA-dplr} 
\captionsetup{labelformat=empty,labelsep=none}
\end{figure} 

\subsection{Relation to DPLR}
\label{sec:related_dplr}

(Gated) DeltaNet can be generalized to a more expressive \emph{Diagonal-Plus-Low-Rank} (DPLR) structure, defined as $\mathbf{D} - \bm{a}_t \bm{b}_t^\top$.
This structure was also explored in models such as S4~\citep{gu-2022-efficiently}, which employed a static DPLR formulation as the state transition matrix.
During computation, this matrix is typically jointly diagonalized into the complex plane, thereby restricting its expressiveness to diagonal transformations~\citep{merrill2024illusion}.

While the DPLR structure introduces richer model interactions and can potentially enhance recall through its key–value update rule, it also suffers from a notable limitation: high computational cost and poor parallelizability.
These drawbacks make DPLR inherently slower in large-scale or real-time scenarios, where maintaining parameter efficiency becomes a crucial design challenge.

To address this issue, KDA introduces a constrained variant of DPLR, where Eq.~\ref{eq:recurrent_KDA} can be rewritten as
$\mathbf{S}_t = \left(
    \brickred{\operatorname{Diag}\!\left(\bm{\alpha}_t \right)} 
    - \beta_t \bm{k}_t \bm{k}_t^{\top} 
      \brickred{\operatorname{Diag}\!\left(\bm{\alpha}_t \right)}
\right)\mathbf{S}_{t-1} 
+ \beta_t \bm{k}_t \bm{v}_t^{\top}
$
with the correspondence between the two given by:
\[
\mathbf{S}_t = (\mathbf{D} - \bm{a}_t \bm{b}_t^{\top}) \mathbf{S}_{t-1} + \bm{k}_t \bm{v}_t^{\top}, \mathrm{s.t.},\;\; \mathbf{D} = \brickred{\operatorname{Diag}\!\left(\bm{\alpha}_t \right)}, 
\bm{a}_t = \beta_t \bm{k}_t,
\bm{b}_t = \bm{k}_t \odot\brickred{\bm{\alpha}_t}.
\]
Furthermore, by sharing $\brickred{\bm{\alpha}_t}$, we can factor it out as in Eq.~\ref{eq:recurrent_KDA}, enabling a fine-grained multiplicative decay over $\mathbf{S}_t$ in a manner similar to GLA~\citep{yang-etal-2024-gla}, followed by a Householder-style transformation like DeltaNet~\citep{schlag-2021-deltanet,yang-2024-parallelizing} for efficient state updating.
We provide a side-by-side comparison of the chunkwise PyTorch-style pseudocode implementations for DPLR and KDA in Listing~\ref{listing:dplr} and Listing~\ref{listing:KDA}.%
The key improvements are highlighted below:
\begin{itemize}[leftmargin=12pt]
    \item Listing~\ref{listing:dplr} \colorbox{red!10}{Line 13-16} vs., Listing~\ref{listing:KDA} \colorbox{green!10}{Line 14-15}: the reciprocal of the cumulative decay term $1/\Gamma$ in chunkwise form (Eq.~\ref{eq:gdn-o}) can introduce numerical instability. While we can resolve this issue by secondary chunking \citep{yang-2024-fla}, it incurs additional computation and I/O overhead. By fixing $\bm{a}=\bm{b} = \bm{k}$ in the DPLR formulation, KDA removes the need for two secondary chunking steps, substantially reducing redundant operations and improving overall efficiency.
    \item Listing~\ref{listing:dplr} \colorbox{red!10}{Line 25-27,31-32} vs., Listing~\ref{listing:KDA} \colorbox{green!10}{Line 26,29}: KDA further eliminates roughly three matrix multiplications during inter-chunk and output computation, leading to significant kernel-level acceleration.
\end{itemize}
We further benchmark the kernel speed in Fig.~\ref{fig:kernel}, showing that KDA achieves nearly $2\times$ the speed of DPLR for sequence lengths up to $64\text{k}$.

\subsection{Complexity Analysis}
\paragraph{Training flops}
We maintain a similar number of parameters in Kimi Linear as in the full attention MLA. The linear projection calculation remains identical to that of the global attention layer. The key distinction lies in the FLOPs associated with attention computation. For simplicity, we focus on non-variable length scenarios. Based on the implementation of the gated rule kernel, the theoretical FLOPs for a single attention head with headdim $d_h$ and a fixed chunk size $C = 64$ in the gated delta rule \citep{wang-deltanet} (per sequence of length $T$) are as follows:
\begin{align}
\mathrm{FLOPs}_{\text{KDA}}(T; C, d_h)
&= 6 T d_h^2 + 3 T C d_h + T C^2. \label{eq:gdn}
\end{align}
For full (global) attention, the dominant term per head is
\begin{equation}
\mathrm{FLOPs}_{\text{Attn}}(T; d_h) \;=\; 2 T^2 d_h. \label{eq:attn}
\end{equation}

\subparagraph{Inference strategy and cost}
The inference strategy in Kimi Linear employs a hybrid approach to optimize both computational and I/O efficiency. During the prefill phase, the model utilizes a FLOP-intensive chunk kernel (see \S~\ref{sec:kda:chunk}), while switching to the more efficient recurrent kernel (Eq. \ref{eq:KDA-recurrent}) for autoregressive generation.
A key advantage of the Linear KDA is its ability to maintain a fixed-sized state ($d_k \times d_v$ per head, with $d_k = d_v = 128$) regardless of sequence length.
For our hybrid model, as sequence length increases, the I/O-bounded decoding time approaches a maximum hybrid efficiency ratio of 3:1 compared to full attention. This trend is reflected in Fig.~\ref{fig:decoding}, where Kimi Linear achieves a $2.3\times$ speedup at a 1M token context. Additionally, by eliminating the need for a large, linear-scaling KV cache, Kimi Linear is able to reallocate memory resources to support larger batch sizes, enhancing overall throughput. In long-context scenarios (up to 1M tokens), this memory efficiency results in a theoretical decoding speedup of up to $6.3\times$ (see Fig.~\ref{fig:decoding-best}).

\section{Related Works}

\subsection{Efficient Subquadratic Attention}
The quadratic time complexity of the standard self-attention mechanism \citep{vaswani-2017-attention} remains a fundamental bottleneck for processing long contexts in Transformer-based models. This limitation has become increasingly critical as large language models (LLMs) are now expected to handle million-token sequences for tasks such as agentic tool use and repository-level code analysis \citep{deepseekaiv3,kimi2025k2}. To overcome this challenge, a substantial body of research has explored more efficient attention mechanisms \citep{sun2025efficient,sun2025speedwinssurveyefficient}, which can broadly be categorized into two main directions: (1) linear attention, and (2) sparse attention.

\begin{table}[t]
    \centering
    \small
    \caption{
        An overview of different attention mechanisms through the lens of state updating rules and their learning objective under the TTT framework \citep{sun-2024-learning}.
        We ignore all normalizer terms and activation/kernel functions for brevity.
    }
    \renewcommand{\arraystretch}{1.8}
    \begin{adjustbox}{width=1\columnwidth, center}
        \renewcommand{\multirowsetup}{\centering}
        \setlength{\tabcolsep}{18pt}
        \begin{threeparttable}
            \begin{tabular}{r ll}
                \toprule
                                                                                                                                                                                                                                                                                                                & Objective $\mathcal{L}$                                                                                                                                                                                   & Update rule $\mathbf{S}_t = \mathbf{S}_{t-1} - \nabla_{\mathbf{S}_{t-1}}\mathcal{L}$                                         \\
                \midrule
                LA \citep{katharopoulos-2020-transformers}                                                                                                                                                                                                                                                      &
                $-\left\langle\mathbf{S}_{t-1}^\top \boldsymbol{k}_t, \boldsymbol{v}_t\right\rangle$                                                                                                                                                                                                            & $\mathbf{S}_t = \mathbf{S}_{t-1}+\boldsymbol{k}_t\boldsymbol{v}_t^\top$                                                                                                                                                                                                                                                                  \\
                RetNet \citep{sun-2023-retnet}                                                                                                                                                                                                                                                                  &
                $-{\beta_t}\left\langle\mathbf{S}_{t-1}^\top \boldsymbol{k}_t, \boldsymbol{v}_t\right\rangle+\frac{1}{2}\left\|\sqrt{\brickred{1-\mathbf{\alpha}}}~\mathbf{S}_{t-1}\right\|_F^2$                                                                                                                & $\mathbf{S}_t = \brickred{\alpha}\mathbf{S}_{t-1}+{\beta_t}\boldsymbol{k}_t\boldsymbol{v}_t^\top$                                                                                                                                                                                                                                        \\
                Mamba2 \citep{mamba2}                                                                                                                                                                                                                                                                           &
                $-{\beta_t}\left\langle\mathbf{S}_{t-1}^\top \boldsymbol{k}_t, \boldsymbol{v}_t\right\rangle+\frac{1}{2}\left\|\sqrt{\brickred{1-\mathbf{\alpha}_t}}\mathbf{S}_{t-1}\right\|_F^2$                                                                                                               & $\mathbf{S}_t = \brickred{\alpha_t}\mathbf{S}_{t-1}+{\beta_t}\boldsymbol{k}_t\boldsymbol{v}_t^\top$                                                                                                                                                                                                                                      \\
                GLA \citep{yang-etal-2024-gla}                                                                                                                                                                                                                                                                  & $-\left\langle\mathbf{S}_{t-1}^\top \boldsymbol{k}_t, \boldsymbol{v}_t\right\rangle+\frac{1}{2}\left\|\sqrt{\brickred{\operatorname{Diag}\left(\bm{1}-\bm{\alpha}_t\right)}}\mathbf{S}_{t-1}\right\|_F^2$ & $\mathbf{S}_t = \brickred{\operatorname{Diag}(\boldsymbol{\alpha}_t)}\mathbf{S}_{t-1}+\boldsymbol{k}_t\boldsymbol{v}_t^\top$ \\
                HGRN2 \citep{qin-2024-hgrn2}                                                                                                                                                                                                                                                                    &
                $-\left\langle\mathbf{S}_{t-1}^\top \brickred{(\mathbf{1-\boldsymbol{\alpha}}_t)}, \boldsymbol{v}_t\right\rangle+\frac{1}{2}\left\|\sqrt{\brickred{\operatorname{Diag}\left(\bm{1}-\bm{\alpha}_t\right)}}\mathbf{S}_{t-1}\right\|_F^2$                                                          &
                $\mathbf{S}_t = \brickred{\operatorname{Diag}(\boldsymbol{\alpha}_t)}\mathbf{S}_{t-1}+\brickred{(\mathbf{\bm{1}-\boldsymbol{\alpha}}_t)}\boldsymbol{v}_t^\top$                                                                                                                                                                                                                                                                                                                                                                                                                                                                             \\
                \midrule

                                Longhorn \citep{liu-2024-longhorn}                                                                                                                                                                                                                                                              &
                $\frac{1}{2}\left\|\mathbf{S}_{t-1}^\top \boldsymbol{k}_t-\boldsymbol{v}_t \right\|^2_{\operatorname{Diag}(\bm{\beta}_t)} $                                                                                                                                                                     & $\mathbf{S}_t = {\left(\mathbf{I}
                    -\frac{\bm{\beta}_t}{\bm{1}+\bm{\beta}_t\bm{k}_t^\intercal\bm{k}_t}\bm{k}_t\bm{k}_t^\intercal\right)}\mathbf{S}_{t-1}+{\beta_t}\bm{k}_t\bm{v}_t^\intercal$
                \\

                Comba \citep{hu2025comba}                                                                                                                                                                                                                                                                     &
                $ \frac{{\beta_t}}{2}\left\|\mathbf{S}_{t-1}^\top \boldsymbol{k}_t -\boldsymbol{v}_t\right\|^2+\frac{1}{2}\left\|\sqrt{\brickred{1-\mathbf{\alpha}_t}}\mathbf{S}_{t-1}\right\|_F^2$                                                                                                             &
                $\mathbf{S}_t = \left(\brickred{\alpha_t}
                -\beta_t\boldsymbol{k}_t\hat{\boldsymbol{k}}_t^\top\right)\mathbf{S}_{t-1}+{\beta_t}\boldsymbol{k}_t\boldsymbol{v}_t^\top$                                                                                                                                                                         \\

                RWKV7 \citep{peng-2025-rwkv7}                                                                                                                                                                                                                                                                   &
                $ \frac{1}{2}\left\|\mathbf{S}_{t-1} ^\top\tilde{\boldsymbol{k}}_t-\boldsymbol{v}_t\right\|^2+\frac{1}{2}\left\|\sqrt{\brickred{\operatorname{Diag}\left(\bm{1}-\bm{\alpha}_t\right)}}\mathbf{S}_{t-1}\right\|_F^2 $                                                         &
                $\mathbf{S}_t = \left(\brickred{\operatorname{Diag}\left(\boldsymbol{\alpha}_t\right)}-\left(\bm{b}_s\odot\hat{\bm{k}}_s\right)\hat{\boldsymbol{k}}_t^\top\right)\mathbf{S}_{t-1}+{\boldsymbol{k}}_t\boldsymbol{v}_t^\top$                                                                                                                                                                                                                                                                                                                                                                                                                      \\

                GDN \citep{yang-2025-gdn}                                                                              &
                $ \frac{{\beta_t}}{2}\left\|\brickred{\tilde{\mathbf{S}}_{t-1}^\top} \boldsymbol{k}_t-\boldsymbol{v}_t\right\|^2 $                                                             &
                $\mathbf{S}_t = {\left(\mathbf{I}-{\beta}_t\boldsymbol{k}_t\boldsymbol{k}_t^\top\right)\brickred{\alpha_t}}\mathbf{S}_{t-1}+{\beta}_t\boldsymbol{k}_t\boldsymbol{v}_t^\top$                                                                       \\

                \rowcolor{gray!20} \textbf{KDA} (ours)                                                                                                                                                                                                                                                         &
                $ \frac{\beta_t}{2}\left\|\brickred{\tilde{\mathbf{S}}_{t-1}^\top} {\boldsymbol{k}_t}-\boldsymbol{v}_t\right\|^2$ &
                $\mathbf{S}_t = \left(\mathbf{I}
                -\beta_t\boldsymbol{k}_t\boldsymbol{k}_t^\top\right)\brickred{\operatorname{Diag}\left(\boldsymbol{\alpha}_t\right)}\mathbf{S}_{t-1}+{\beta_t}\boldsymbol{k}_t\boldsymbol{v}_t^\top$                                                                                                                                                                                                                                                                                                                                                                                                                                                       \\

                \bottomrule
            \end{tabular}
            \begin{tablenotes}
            \footnotesize
              {\item For GDN and KDA, the update can be viewed as performing an Stochastic Gradient Descent(SGD) process on the decayed state $\brickred{\tilde{\mathbf{S}}}$, that is, $\mathbf{S}_t = \brickred{\tilde{\mathbf{S}}_{t-1}} - \nabla_{\brickred{\tilde{\mathbf{S}}_{t-1}}}\mathcal{L}$, where $\brickred{\tilde{\mathbf{S}}_{t-1}}$ is decayed by scalar or fine-grained gate.}
            \end{tablenotes}
        \end{threeparttable}
    \end{adjustbox}
    \label{tab:KDA-obj}
\end{table}
\paragraph{Linear Attention}
reformulates the quadratic attention map into kernelized feature interactions, replacing the softmax with a positive feature map so that attention can be computed through two associative matrix products \citep{katharopoulos-2020-transformers}. 
This eliminates the explicit $\mathcal{O}(T^2)$ similarity matrix and enables linear-time computation with respect to sequence length.
Subsequent work strengthens the vanilla linear attention significantly through more refined memory control, shifting from data-independent ``decay'' \citep{sun-2023-retnet,qin-2024-transnormerllm} to more adaptive, data-dependent mechanisms \citep{gu-2023-mamba,sun-2024-yoco}, and refining the decay granularity from coarse headwise \citep{mamba2} to precise, channel-wise decay.
GLA generalizes these approaches with diagonal, channel-wise gates that balance expressiveness and efficiency while retaining chunk-wise parallelism \citep{yang-2024-fla,yang-etal-2024-gla}. 
Table~\ref{tab:KDA-obj} summarizes the corresponding update rules. 
Collectively, these methods cast attention as a compact recurrent memory updated with parallel prefix-scan operators and fused matrix multiplies, aligning well with modern accelerators \citep{hua-etal-2022-gau}.

A complementary view connects linear attention to \emph{fast-weight} memory \citep{schlag-2021-deltanet}: the state is a low-capacity associative table updated online by Hebbian-like rules \citep{munkhdalai-2019-metalearned}, while slow weights amortize when to store, update, or forget \citep{munkhdalai2018metalearninghebbianfastweights}. 

In Table \ref{tab:KDA-obj}, we provide a summary of the existing efficient token mixing methods, comparing them from the perspectives of state update mechanisms and optimization objectives.

From this perspective, gating and decay serve as learnable criteria that mitigate interference and stabilize optimization \citep{sun-2024-learning}. 
Despite these advances, linear attention still lags full attention on exact copying and fine-grained selection in extreme long-context retrieval. 
This motivates hybrid designs (interleaving linear and full attention) and more structured updates. 
In particular, the gated delta rule used by GDN/KDA introduces rank-1 corrective updates to the fast-weight state, improving targeted retention while remaining parallelizable at the operator level \citep{yang-2024-parallelizing}.

\paragraph{Linear Attention with Gating Mechanism}
The vanilla Linear Attention \citep{katharopoulos-2020-transformers} is known to lack the selection mechanism inherent in softmax attention \citep{vaswani-2017-attention}, falling short in expressiveness. 
To address this, Gated Linear Attention models have emerged as memory-efficient and parallelizable alternatives \citep{yang-2024-fla,yang-etal-2024-gla,gu-2023-mamba}. Instead of storing an ever-expanding KV cache, these models employ a fixed-size matrix-valued state and learnable gates to selectively retain and forget information.  
This design achieves expressive power comparable to softmax attention \citep{merrill-sabharwal-2023-parallelism, zhong2025understanding,merrill2024illusion} while maintaining constant time and memory complexity during inference time. 
The general recurrent formulation of such models for memory update $\mathbf{S}_t \in \mathbb{R}^{d_k \times d_v}$ can be expressed as:
\begin{equation}
    \mathbf{S}_{t}=\brickred{\mathbf{A}_t}\mathbf{S}_{t-1} +  \bm{k}_t\bm{v}_t^\top,\quad\bm{o}_{t}=\mathbf{S}_t^\top\bm{q}_t.
    \label{eq:linear-attn}
\end{equation}
The primary distinction among various gated linear attention mechanisms lies in the parameterization of the forget gate $\brickred{\mathbf{A}_t}$, as summarized in Table~\ref{tab:KDA-obj}.
For instance, RetNet \citep{sun-2023-retnet} uses a data-independent scalar decay $\brickred{\alpha}$, and Mamba2 \citep{mamba2} employs a data-dependent scalar $\brickred{\alpha_t}$.
Specifically,  GLA \citep{yang-etal-2024-gla} utilized a diagonalized fine-grained matrix $\brickred{\operatorname{Diag}(\bm{\alpha}_t)}\in \mathbb{R}^{d_k\times d_k}$, offering an effective trade-off between efficiency and performance. Other variants are displayed in Table~\ref{tab:KDA-obj}.

\paragraph{Sparse Attention}

A separate body of work reduces the quadratic complexity of standard attention by exploiting its inherent sparsity, approximating the full attention score by performing the computation on a strategically selected subset of tokens.
The central challenge lies in identifying this subset effectively without degrading model performance. 
Early methods often utilized efficient, training-free static patterns, such as sliding and dilated windows \citep{ding2023longnetscalingtransformers1000000000,gu2025attentionsinkemergeslanguage,xiao2023efficient}, or fixed patterns \citep{zaheer2020big,guo2019star}, but their rigid structure often compromises model accuracy. 
More advanced methods determine the important positions based on the context, such as clustering \citep{kitaev2020reformer,wu2022memorizing} and lightweight routing mechanisms \citep{fu2024moa,pikekos2025mixture,ainslie2023colt5,bertsch2023unlimiformer}, but this dynamic selection process introduces a computational overhead that can prevent them from achieving their full theoretical speedup without dedicated kernel acceleration \citep{dong2024flexattentionprogrammingmodel}. Some models further introduce training-free sparsification during the inference stage \citep{xiao2023efficient,xu2025xattention}.

Recent approaches to sparse attention have begun to prioritize hardware co-design, as exemplified by NSA \citep{yuan2025nativesparseattentionhardwarealigned,minicpmteam2025minicpm4ultraefficientllmsend} and MoBA \citep{lu2025mobamixtureblockattention}, which both move from token-level to chunk-level selection. 
In NSA, each query dynamically selects chunks based on scores produced by an MLP. The method’s efficiency relies on its use of Grouped-Query Attention (GQA) \citep{touvron-2023-llama} with a large head count (typically a multiple of 16), a configuration specifically designed to accelerate computation through highly parallelized tensor–matrix multiplications.
Similarly, MoBA performs top-$k$ chunk selection, but leverages log-sum-exp (LSE) scores computed efficiently via flash-attention kernels \citep{dao-2022-flashattn}. In contrast to NSA and MoBA, the recently proposed DeepSeek-V3.2-Exp Attention (DSA) \citep{deepseekai2024deepseekv32} revives token-level sparsity, maintaining efficiency through a learnable full-attention indexer implemented with low-precision fp8 and a small head dimension for token selection.

\paragraph{Discussion} Linear attention and sparse attention represent two distinct pathways toward efficient long-context modeling. Sparse attention tends to retrieve fine-grained historical information more effectively, but this advantage comes at the cost of storing the entire KV cache for token selection, making it less efficient than linear attention models that maintain a constant state. Moreover, sparse attention performs only information selection, and its theoretical expressive upper bound remains that of full attention. In contrast, linear attention, grounded in the principle of “compression as intelligence”, enables generalization with a fixed-size state and, when combined with the Delta learning rule, can achieve theoretically stronger expressive capacity. Although linear attentions have traditionally been criticized for weak retrieval ability, this limitation can be mitigated through state expansion \citep{du2025mom,guo2025log,yau2025sequential,hu2024attractor} or related techniques. Nevertheless, despite these advantages, linear attention remains limited by current hardware implementations and the absence of optimized inference infrastructure. Our work overcomes these limitations with Kimi Linear, a powerful model integrated with vLLM for efficient inference. Our proposed KDA delivers competitive performance compared to the full-attention baseline (Table~\ref{tab:pretrain}) and achieves over a $2\times$ decoding speedup at the one-million-token context (Figure~\ref{fig:decoding}). Despite their distinct approaches to efficient long-context modeling, linear attention and sparse attention are not mutually exclusive. Future work could explore hybrid models that integrate the strengths of both, leveraging the compression and generalization capabilities of linear attention with the fine-grained retrieval advantages of sparse attention to further enhance model performance and efficiency.

\subsection{Hybrid Model}
\label{sec:hybrid}

Despite efficiency, pure Linear Attention still struggle with precise memory retrieval and exact copying \citep{jelassi-2024-repeat,wen2024rnns}
This deficiency hinders their adoption in industrial-scale LLMs where robust long-context recall (e.g., beyond 1M tokens) and reliable tool-use over extensive code repositories are critical \citep{kimi2025k2}. Recent work shows that Linear Attention and full attention can effectively complement each other, leading to various hybrid designs.

\paragraph{Intra-layer hybrid}
One category of hybrid architectures is the intra-layer hybrid, which adaptively fuses the outputs of different mechanisms within each layer.
A common implementation fuses outputs from heterogeneous heads within each layer, such as combining standard attention with state space models (SSMs) \citep{dong2024hymba,li2025transmamba}.
In contrast, sequence-level approaches apply distinct mechanisms to different parts of the input. For example, some use linear attention for past context and SWA for recent tokens \citep{zhang2024lolcats,lan2025liger,munkhdalai2024leave}, while NHA \citep{du2025native} compresses the history with GSA \citep{zhang2024gated} and combines it with local sliding window context to emulate a standard attention operation.

\paragraph{Inter-layer Hybrid} 
A key drawback of the intra-layer hybrid is the increased system complexity and inference overhead. 
The heterogeneous mechanisms require separate computational paths, complicating optimizations like distributed parallelism. To mitigate this challenge, inter-layer hybrids have become a more widely adopted and practical strategy in LLMs \citep{minimax2025minimax01,lieber2024jamba,team2025hunyuan}.
This approach involves stacking distinct layer types, such as full attention and a linear alternative, in a predefined ratio.
Building on this paradigm, we implement a simple yet effective strategy: interleaving linear and full attention layers at a fixed 3:1 ratio (see \S~\ref{sec:ablation} for ablations). This regular, repeating structure simplifies KV cache management and integrates seamlessly with standard optimizations.
For the linear component of our hybrid, we deviate from the common practice of using Mamba2 \citep{mamba2}. Instead, we employ KDA, as we found it yields superior overall performance, particularly in retrieval and copying abilities.

\paragraph{Discussion}
Recent work indicates that hybrid models can be sensitive to adjustments in the RoPE base frequency, a vulnerability that complicates context window extension \citep{zuo2025falconh1familyhybridheadlanguage}. This sensitivity can hinder the model's ability to extrapolate to longer sequences.
To address this challenge, recent models have trended towards solutions that incorporate No Position Embeddings (NoPE). Falcon-H \citep{zuo2025falconh1familyhybridheadlanguage}, for example, uses an unconventionally high base frequency (e.g., $b \approx 10^{11}$) to push its positional encoding to a near-NoPE state.
Architecturally, SwanGPT \citep{puvvada2025swangpt} interleaves RoPE-based layers with NoPE-based full attention layers. Aligning with this direction, we found that hybridizing our KDA layers with NoPE full attention is also a highly effective strategy, facilitating straightforward context window extension.

\section*{Conclusion}

We introduce Kimi Linear, a hybrid linear attention architecture designed to meet the efficiency demands of agentic intelligence and test-time scaling without sacrificing quality. Central to Kimi Linear is Kimi Delta Attention (KDA), an advanced linear attention module with a channel-wise gating mechanism that enhances memory control and enables RNN-style models in hybrid architectures. By interleaving KDA with global attention in a 3:1 ratio, Kimi Linear reduces memory usage by up to 75\%, while achieving up to 6.3$\times$ higher decoding throughput and outperforming full-attention baselines. Our approach provides a scalable, efficient solution for large language models, with open-source KDA kernels and pre-trained checkpoints facilitating further research.

\newpage
\printbibliography[title={References}]

\newpage
\appendix
\section{Contributions}
The authors are listed in order of the significance of their contributions, with those in project leadership roles appearing last. The project is developed at Moonshot AI, with several external collaborators that are marked with \#. Names marked with an asterisk (*) indicate people who are no longer part of our team. 
\begin{multicols}{1}
Yu Zhang$^{1}$\\
Zongyu Lin$^{*}$\\
Xingcheng Yao\\
Jiaxi Hu$^2$\\
Fanqing Meng\\
Chengyin Liu\\
Xin Men\\
Songlin Yang$^\#$$^3$\\
Zhiyuan Li\\
Wentao Li\\
Enzhe Lu\\
Weizhou Liu\\
Yanru Chen\\
Weixin Xu\\
Longhui Yu\\
Yejie Wang\\
Yu Fan\\
Longguang Zhong\\
Enming Yuan\\
Dehao Zhang\\
Yizhi Zhang\\
T.Y. Liu\\
Haiming Wang\\
Shengjun Fang\\
Weiran He\\
Shaowei Liu\\
Yiwei Li\\
Jianlin Su\\
Jiezhong Qiu$^4$\\
Bo Pang\\
Junjie Yan\\
Zhejun Jiang\\
Weixiao Huang\\
Bohong Yin\\
Jiacheng You\\
Chu Wei\\
Zhengtao Wang\\
Chao Hong\\
Yutian Chen\\
Guanduo Chen\\
Yucheng Wang\\
Huabin Zheng\\
Feng Wang\\
Yibo Liu\\
Mengnan Dong\\
Zheng Zhang\\
Siyuan Pan\\
Wenhao Wu\\
Yuhao Wu\\
Longyu Guan\\
Jiawen Tao\\
Guohong Fu$^\#$$^1$\\
Xinran Xu\\
Yuzhi Wang\\
Guokun Lai\\
Yuxin Wu\\
Xinyu Zhou\\
Zhilin Yang\\
Yulun Du\\
\end{multicols}

$^1$ Soochow University, China

$^2$ The Hong Kong University of Science and Technology (Guangzhou)

$^3$ Massachusetts Institute of Technology

$^4$ Hangzhou Institute of Medicine, CAS

\newpage

\section{Derivations for Chunkwise Parallelism of KDA}

We first recall the recurrent form of KDA:
\begin{equation*}
    \begin{aligned}
        \mathbf{S}_{[t]}^r & = \underbrace{\left(\prod_{i=1}^r \left(\mathbf{I} - \beta_{[t]}^i \boldsymbol{k}_{[t]}^i \boldsymbol{k}_{[t]}^{i\top}\right) \brickred{\operatorname{Diag}(\boldsymbol{\alpha}_{[t]}^i)}\right)}_{:= \mathbf{P}_{[t]}^r} \cdot\mathbf{S}_{[t]}^{0} + \underbrace{\sum_{i=1}^{r} \left(\prod_{j=i+1}^r \left(\mathbf{I} - \beta_{[t]}^j \boldsymbol{k}_{[t]}^j \boldsymbol{k}_{[t]}^{j\top}\right)\brickred{\operatorname{Diag}(\boldsymbol{\alpha}_{[t]}^j)}\right)\cdot\beta_{[t]}^i \boldsymbol{k}_{[t]}^i\boldsymbol{v}_{[t]}^{i\top}}_{:=\mathbf{H}_{[t]}^r} \\
                     & = \mathbf{P}_{[t]}^r \cdot \mathbf{S}_{[t]}^0 + \mathbf{H}_{[t]}^r
    \end{aligned}
    \label{eq:state_update}
\end{equation*}
Our goal is to transform $\mathbf{P}_{[t]}^r$ and $\mathbf{H}_{[t]}^r$ into matrix forms suitable for parallel computation.

We show that $\mathbf{P}_{[t]}^r$, which involves the cumulative product of generalized Householder matrices, can be optimized using the classic WY representation.

\begin{proposition}
    The matrix $\mathbf{P}_{[t]}^r$ can be expressed as:
    \begin{equation}
        \mathbf{P}_{[t]}^r = \brickred{\operatorname{Diag}(\boldsymbol{\gamma}_{[t]}^r)} - \sum_{i=1}^{r} \brickred{\operatorname{Diag}(\boldsymbol{\gamma}_{[t]}^{i\rightarrow r})} \boldsymbol{k}_{[t]}^i \boldsymbol{w}_{[t]}^{i\top}
        \label{eq:P_wy}
    \end{equation}
    where the auxiliary vector $\boldsymbol{w}_{[t]}^r \in \mathbb{R}^{d_k}$ is computed via the following recurrence relation:
    \begin{equation}
        \boldsymbol{w}_{[t]}^r = \beta_{[t]}^r \left( \brickred{\operatorname{Diag}(\boldsymbol{\gamma}_{[t]}^r)} \boldsymbol{k}_{[t]}^r - \sum_{i=1}^{r-1} \boldsymbol{w}_{[t]}^i\left( \boldsymbol{k}_{[t]}^{i\top}\brickred{\operatorname{Diag}\left(\boldsymbol{\gamma}_{[t]}^{i\rightarrow r} \right)}\boldsymbol{k}_{[t]}^r \right)  \right)
        \label{eq:w_recursion}
    \end{equation}
\end{proposition}

\begin{proof}
    We proceed with a proof by mathematical induction.

    \textbf{Inductive Step:} Assume the proposition holds for $r-1$, i.e., $\mathbf{P}_{[t]}^{r-1} = \brickred{\operatorname{Diag}(\boldsymbol{\gamma}_{[t]}^{r-1})} - \sum_{i=1}^{r-1} \brickred{\operatorname{Diag}(\boldsymbol{\gamma}_{[t]}^{i\rightarrow r-1})} \boldsymbol{k}_{[t]}^i \boldsymbol{w}_{[t]}^{i\top}$.
    We now derive:
    \begin{align*}
        \mathbf{P}_{[t]}^r & = \left(\mathbf{I} - \beta_{[t]}^r \boldsymbol{k}_{[t]}^r \boldsymbol{k}_{[t]}^{r\top}\right)\brickred{\brickred{\operatorname{Diag}(\boldsymbol{\alpha}_{[t]}^r)}}\mathbf{P}_{[t]}^{r-1}                                                            \\
                       & = \left(\mathbf{I} - \beta_{[t]}^r \boldsymbol{k}_{[t]}^r \boldsymbol{k}_{[t]}^{r\top}\right)\brickred{\brickred{\operatorname{Diag}(\boldsymbol{\alpha}_{[t]}^r)}}\left(\brickred{\operatorname{Diag}(\boldsymbol{\gamma}_{[t]}^{r-1})} - \sum_{i=1}^{r-1} \brickred{\operatorname{Diag}\left(\boldsymbol{\gamma}_{[t]}^{i\rightarrow r-1}\right)} \boldsymbol{k}_{[t]}^i \boldsymbol{w}_{[t]}^{i\top}\right)                                                                                                                                         \\
                       & = \left(\mathbf{I} - \beta_{[t]}^r \boldsymbol{k}_{[t]}^r \boldsymbol{k}_{[t]}^{r\top}\right)\left(\brickred{\operatorname{Diag}(\boldsymbol{\gamma}_{[t]}^r)} - \sum_{i=1}^{r-1} \brickred{\operatorname{Diag}\left(\boldsymbol{\gamma}_{[t]}^{i\rightarrow r}\right)} \boldsymbol{k}_{[t]}^i \boldsymbol{w}_{[t]}^{i\top}\right)                                                                                                                                         \\
                       & = \brickred{\operatorname{Diag}(\boldsymbol{\gamma}_{[t]}^r)} -\sum_{i=1}^{r-1} \brickred{\operatorname{Diag}\left(\boldsymbol{\gamma}_{[t]}^{i\rightarrow r}\right)} \boldsymbol{k}_{[t]}^i \boldsymbol{w}_{[t]}^{i\top} - \beta_{[t]}^r \boldsymbol{k}_{[t]}^r  \boldsymbol{k}_{[t]}^{r\top} \brickred{\operatorname{Diag}(\boldsymbol{\gamma}_{[t]}^r)} + \beta_{[t]}^r \boldsymbol{k}_{[t]}^r  \boldsymbol{k}_{[t]}^{r\top} \sum_{i=1}^{r-1} \brickred{\operatorname{Diag}\left(\boldsymbol{\gamma}_{[t]}^{i\rightarrow r}\right)} \boldsymbol{k}_{[t]}^i \boldsymbol{w}_{[t]}^{i\top}                                                                                                                                    \\
                       & = \brickred{\operatorname{Diag}(\boldsymbol{\gamma}_{[t]}^r)} - \sum_{i=1}^{r-1} \brickred{\operatorname{Diag}\left(\boldsymbol{\gamma}_{[t]}^{i\rightarrow r}\right)} \boldsymbol{k}_{[t]}^i \boldsymbol{w}_{[t]}^{i\top} - \boldsymbol{k}_{[t]}^r  \left(\beta_{[t]}^r \brickred{\operatorname{Diag}(\boldsymbol{\gamma}_{[t]}^r)}\boldsymbol{k}_{[t]}^r  \right)^\top + \boldsymbol{k}_{[t]}^r  \left(\beta_{[t]}^r \sum_{i=1}^{r-1} \bm{w}^i_{[t]}  \left( \boldsymbol{k}_{[t]}^{i\top} \brickred{\operatorname{Diag}\left(\boldsymbol{\gamma}_{[t]}^{i\rightarrow r}\right)} \boldsymbol{k}_{[t]}^r  \right)  \right)^\top                                                                                                          \\
                       & = \brickred{\operatorname{Diag}(\boldsymbol{\gamma}_{[t]}^r)} - \sum_{i=1}^{r-1} \brickred{\operatorname{Diag}\left(\boldsymbol{\gamma}_{[t]}^{i\rightarrow r}\right)} \boldsymbol{k}_{[t]}^i \boldsymbol{w}_{[t]}^{i\top} - \boldsymbol{k}_{[t]}^r   \underbrace{\left(\beta_{[t]}^r \left( \brickred{\operatorname{Diag}(\boldsymbol{\gamma}_{[t]}^r)}\boldsymbol{k}_{[t]}^r  - \sum_{i=1}^{r-1} \boldsymbol{w}_{[t]}^{i} \left( \boldsymbol{k}_{[t]}^{i\top} \brickred{\operatorname{Diag}\left(\boldsymbol{\gamma}_{[t]}^{i\rightarrow r}\right)} \boldsymbol{k}_{[t]}^r  \right)  \right)\right)^\top}_{\boldsymbol{w}_{[t]}^r}  \\
                       & = \brickred{\operatorname{Diag}(\boldsymbol{\gamma}_{[t]}^r)} - \sum_{i=1}^{r-1} \brickred{\operatorname{Diag}\left(\boldsymbol{\gamma}_{[t]}^{i\rightarrow r}\right)} \boldsymbol{k}_{[t]}^i \boldsymbol{w}_{[t]}^{i\top} - \boldsymbol{k}_{[t]}^r  \boldsymbol{w}_{[t]}^{r\top}                                                                                                                                                             \\
                       & = \brickred{\operatorname{Diag}(\boldsymbol{\gamma}_{[t]}^r)} - \sum_{i=1}^{r} \brickred{\operatorname{Diag}\left(\boldsymbol{\gamma}_{[t]}^{i\rightarrow r}\right)} \boldsymbol{k}_{[t]}^i \boldsymbol{w}_{[t]}^{i\top}
    \end{align*}
    The inductive step holds.
\end{proof}

Similar to $\mathbf{P}_{[t]}^r$, $\mathbf{H}_{[t]}^r$ can also be expressed in a parallelizable form.

\begin{proposition}
    The matrix $\mathbf{H}_{[t]}^r$ can be expressed as:
    \begin{equation}
        \mathbf{H}_{[t]}^r = \sum_{i=1}^{r} \brickred{\operatorname{Diag}\left(\boldsymbol{\gamma}_{[t]}^{i\rightarrow r}\right)} \boldsymbol{k}_{[t]}^i \boldsymbol{u}_{[t]}^{i\top}
        \label{eq:H_wy}
    \end{equation}
    where the auxiliary vector $\boldsymbol{u}_{[t]}^r \in \mathbb{R}^{d_v}$ is computed via the following recurrence relation:
    \begin{equation}
        \boldsymbol{u}_{[t]}^r = \beta_{[t]}^r \left(\boldsymbol{v}_{[t]}^r - \sum_{i=1}^{r-1}\boldsymbol{u}_{[t]}^i \left(\boldsymbol{k}_{[t]}^{i\top} \brickred{\operatorname{Diag}\left(\boldsymbol{\gamma}_{[t]}^{i\rightarrow r}\right)} \boldsymbol{k}_{[t]}^r\right)  \right)
        \label{eq:u_recursion}
    \end{equation}
\end{proposition}

\begin{proof}
    We again use mathematical induction.

    \textbf{Inductive Step:} Assume the proposition holds for $r-1$.
    \begin{align*}
        \mathbf{H}_{[t]}^r & = \left(\mathbf{I} - \beta_{[t]}^r \boldsymbol{k}_{[t]}^r  \boldsymbol{k}_{[t]}^{r\top}\right) \brickred{\brickred{\operatorname{Diag}(\boldsymbol{\alpha}_{[t]}^r)}}\mathbf{H}_{[t]}^{r-1} +  \beta_{[t]}^r \boldsymbol{k}_{[t]}^r  \boldsymbol{v}_{[t]}^{r\top}                                                                                                                                                                                             \\
                       & = \left(\mathbf{I} - \beta_{[t]}^r \boldsymbol{k}_{[t]}^r  \boldsymbol{k}_{[t]}^{r\top}\right) \brickred{\brickred{\operatorname{Diag}(\boldsymbol{\alpha}_{[t]}^r)}} \left(\sum_{i=1}^{r-1} \brickred{\operatorname{Diag}\left(\boldsymbol{\gamma}_{[t]}^{i\rightarrow r-1}\right)} \boldsymbol{k}_{[t]}^i \boldsymbol{u}_{[t]}^{i\top}\right) +  \beta_{[t]}^r \boldsymbol{k}_{[t]}^r  \boldsymbol{v}_{[t]}^{r\top}                                                                                \\
                       & = \left(\mathbf{I} - \beta_{[t]}^r \boldsymbol{k}_{[t]}^r  \boldsymbol{k}_{[t]}^{r\top}\right) \left(\sum_{i=1}^{r-1} \brickred{\operatorname{Diag}\left(\boldsymbol{\gamma}_{[t]}^{i\rightarrow r}\right)} \boldsymbol{k}_{[t]}^i \boldsymbol{u}_{[t]}^{i\top}\right) +  \beta_{[t]}^r \boldsymbol{k}_{[t]}^r  \boldsymbol{v}_{[t]}^{r\top}                                                                                \\
                       & = \sum_{i=1}^{r-1}\brickred{\operatorname{Diag}\left(\boldsymbol{\gamma}_{[t]}^{i\rightarrow r}\right)}{\boldsymbol{k}}_{[t]}^i \boldsymbol{u}_{[t]}^{i\top} - \beta_{[t]}^r \boldsymbol{k}_{[t]}^r  \boldsymbol{k}_{[t]}^{r\top} \sum_{i=1}^{r-1}\brickred{\operatorname{Diag}\left(\boldsymbol{\gamma}_{[t]}^{i\rightarrow r}\right)}\boldsymbol{k}_{[t]}^i \boldsymbol{u}_{[t]}^{i\top} +\beta_{[t]}^r \boldsymbol{k}_{[t]}^r  \boldsymbol{v}_{[t]}^{r\top}                       \\
                       & = \sum_{i=1}^{r-1}\brickred{\operatorname{Diag}\left(\boldsymbol{\gamma}_{[t]}^{i\rightarrow r}\right)}\boldsymbol{k}_{[t]}^i \boldsymbol{u}_{[t]}^{i\top} - \boldsymbol{k}_{[t]}^r  \left( \beta_{[t]}^r \sum_{i=1}^{r-1} \left( \boldsymbol{k}_{[t]}^{r\top} \brickred{\operatorname{Diag}\left(\boldsymbol{\gamma}_{[t]}^{i\rightarrow r}\right)} \boldsymbol{k}_{[t]}^i \right) \boldsymbol{u}_{[t]}^i \right)^\top +  \boldsymbol{k}_{[t]}^r \beta_{[t]}^r \boldsymbol{v}_{[t]}^{r\top} \\
                       & = \sum_{i=1}^{r-1}\brickred{\operatorname{Diag}\left(\boldsymbol{\gamma}_{[t]}^{i\rightarrow r}\right)}\boldsymbol{k}_{[t]}^i \boldsymbol{u}_{[t]}^{i\top} + \boldsymbol{k}_{[t]}^r  \left( \underbrace{\beta_{[t]}^r \left(\boldsymbol{v}_{[t]}^r - \sum_{i=1}^{r-1}\boldsymbol{u}_{[t]}^i\left(\boldsymbol{k}_{[t]}^{i\top}\brickred{\operatorname{Diag}\left(\boldsymbol{\gamma}_{[t]}^{i\rightarrow r}\right)}\boldsymbol{k}_{[t]}^r\right) \right)}_{\boldsymbol{u}_{[t]}^r} \right)^\top                 \\
                       & = \sum_{i=1}^{r-1}\brickred{\operatorname{Diag}\left(\boldsymbol{\gamma}_{[t]}^{i\rightarrow r}\right)}\boldsymbol{k}_{[t]}^i \boldsymbol{u}_{[t]}^{i\top} + \boldsymbol{k}_{[t]}^r  \boldsymbol{u}_{[t]}^{r\top}                                                                                                                                                   \\
                       & = \sum_{i=1}^{r} \brickred{\operatorname{Diag}\left(\boldsymbol{\gamma}_{[t]}^{i\rightarrow r}\right)}\boldsymbol{k}_{[t]}^i \boldsymbol{u}_{[t]}^{i\top}
    \end{align*}
    The inductive step holds.
\end{proof}

\newpage
\section{Pseudo Code for chunkwise KDA}
\renewcommand{\theFancyVerbLine}{\ttfamily\textcolor[rgb]{0.5,0.5,0.5}{\scriptsize\arabic{FancyVerbLine}}}
\begin{longlisting}
\begin{minted}[
    mathescape,
    linenos,
    numbersep=4pt,
    frame=lines,
    fontsize=\small,
    framesep=4mm,
]{python}
def chunk_kda(
    q: torch.Tensor,
    k: torch.Tensor,
    v: torch.Tensor,
    g: torch.Tensor,
    beta: torch.Tensor,
    initial_state: Optional[torch.Tensor] = None,
    chunk_size: int = 64
):
    dtype = v.dtype
    B, T, H, K, V, C = *q.shape, v.shape[-1], chunk_size
    N = T // C

    q, k, v, g, beta = map(
        lambda x: rearrange(x, 'b (n c) h ... -> b h n c ...', c=C).to(torch.float),
        [q, k, v, g, beta]
    )
    q = q * K**-0.5
    g = g.cumsum(-2)
    mask = torch.triu(torch.ones(C, C, dtype=torch.bool, device=q.device), diagonal=0)

    A = torch.zeros(B, H, N, C, C, dtype=torch.float, device=q.device)
    for i in range(C):
        k_i = k[..., i, :]
        g_i = g[..., i:i+1, :]
        A[..., i] = torch.einsum('... c d, ... d -> ... c', k * (g - g_i).exp(), k_i)
    A = A * beta[..., None]
    # matrix inverse by forward substitution
    A = -A.masked_fill(mask, 0)
    for i in range(1, C):
        A[..., i, :i] = A[..., i, :i].clone() + (A[..., i, :, None].clone() * A[..., :, :i].clone()).sum(-2)
    A = (A + torch.eye(C, dtype=torch.float, device=q.device)) * beta[..., None, :]

    w = A @ (g.exp() * k)
    u = A @ v

    S = k.new_zeros(B, H, K, V)
    if initial_state is not None:
        S += initial_state
    o = torch.zeros_like(v)
    # strictly lower triangular
    mask = torch.triu(torch.ones(C, C, dtype=torch.bool, device=q.device), diagonal=1)
    for i in range(0, N):
        # [B, H, C, ...]
        q_i, k_i, u_i, g_i, w_i = q[:, :, i], k[:, :, i], u[:, :, i], g[:, :, i], w[:, :, i]
        A = torch.zeros(B, H, C, C, dtype=torch.float, device=q.device)
        # secondary chunking for numerical stability
        for j in range(C):
            k_j = k[:, :, i, j]
            g_j = g[:, :, i, j:j+1, :]
            A[..., j] = torch.einsum('... c d, ... d -> ... c', q_i * (g_i - g_j).exp(), k_j)
        A = A.masked_fill(mask, 0)
        v_i = u_i - w_i @ S
        o[:, :, i] = (q_i * g_i.exp()) @ S + A @ v_i
        S = S * rearrange(g_i[:, :, -1].exp(), 'b h k -> b h k 1')
        S += rearrange((g_i[:, :, -1:] - g_i).exp() * k_i, 'b h c k -> b h k c') @ v_i
    return rearrange(o, 'b h n c d -> b (n c) h d').to(dtype)

\end{minted}
\caption{
    Pseudo PyTorch-style code snippet for KDA chunked form.
}
\label{listing:code}
\end{longlisting}
 
\section{Kimi Linear\string@5.7T results}
\label{appendix:results}

Following Moonlight, we also trained Kimi Linear with an extended 5.7T token dataset to demonstrate its effectiveness. With $3\times$ sparsity and a new attention architecture design, Kimi Linear consistently outperforms Moonlight across nearly all benchmarks, underscoring the efficacy of the new architecture. The results are shown in Table~\ref{tab:pretrain-eval} for base model and Table~\ref{tab:instruct-eval} for instruction tuned model. 
Moonlight-Instruct was not evaluated (``-'') on tasks exceeding its 8K context limit.

Kimi Linear\string@5.7T obtains a score of 94.8 on RULER at 1M context length. This long context performance reinforces that Kimi Linear is a promising alternative to full-attention architectures, delivering comparable or superior results while potentially offering more efficient resource utilization.

\begin{table}[h]
    \centering
    \footnotesize
    \renewcommand{\arraystretch}{1.1}
    \setlength{\tabcolsep}{5pt}
    \caption{Performance of Kimi-Linear-Base and Moonlight-Base across diverse tasks.
    }
    \label{tab:pretrain-eval}
    \begin{tabular}{@{}r l | c | >{\columncolor{gray!20}}c c }
        \toprule
         & \textbf{Benchmark}          & \textbf{\#Shots} & {\textbf{Kimi-Linear-Base}} & \textbf{Moonlight-Base}  \\
        \midrule
         & Architecture                & -                & MoE                         & MoE                                  \\
         & \# Activated Params         & -                & 3B                          & 3B                                    \\
         & \# Total Params             & -                & 48B                         & 16B                                      \\
         & Trained Tokens              & -                & 5.7T                        & 5.7T                                \\
        \midrule
        \multirow{6}{*}{\textit{General}}
         & TriviaQA                    & 5-shots          & 75.2               & 66.2                                     \\
         & SimpleQA                    & 5-shots          & 10.1               & \white{0}5.6                    \\
         & MMLU-Pro                    & 5-shots          & 54.8                        & 42.4                                \\
         & MMLU-redux                  & 5-shots          & 79.7                        & 73.8                                   \\
         & WinoGrande                  & 5-shots          & 81.5               & 74.6                                   \\
         & GPQA-Diamond (avg@8) & 5-shots          & 40.4                        & 35.2                                  \\
        \midrule
        \multirow{4}{*}{\textit{Math}}
         & MATH                        & 4-shots          & 58.5                        & 45.3                                 \\
         & GSM8k                       & 8-shots          & 86.3                        & 77.2                                \\
         & GSM8k-platinum              & 8-shots          & 89.6                        & 79.4                                  \\
         & CMATH                       & 6-shots          & 85.5               & 79.6                                   \\
        \midrule
        \multirow{4}{*}{\textit{Code}}
         & CRUXEval-I-cot              & 0-shots          & 61.0                        & 45.9                                    \\
         & CRUXEval-O-cot              & 0-shots          & 67.0                        & 46.6                               \\
         & LiveCodeBench (v6)           & 1-shots          & 20.0                        & 14.3                               \\
         & EvalPlus                    & -                & 64.9                        & 50.3                               \\
        \midrule
        \multirow{2}{*}{\textit{Chinese}}
         & C-Eval                      & 5-shots          & 83.3                        & 77.6                               \\
         & CSimpleQA                   & 5-shots          & 53.5               & 34.7                              \\
        \bottomrule
    \end{tabular}
\end{table}

\begin{table}[h]
    \centering
    \footnotesize
    \setlength{\tabcolsep}{5pt}
    \caption{Performance of Kimi-Linear-Instruct and Moonlight-Instruct across diverse tasks.
    }
    \label{tab:instruct-eval}
    \begin{tabular}{@{}r l | >{\columncolor{gray!20}}c c }
        \toprule
         & \textbf{Benchmark}         & {\textbf{Kimi-Linear-Instruct}} & {\textbf{Moonlight-Instruct}} \\
        \midrule
         & Architecture                                & MoE                             & MoE                                               \\
         & \# Activated Params                         & 3B                              & 3B                                                  \\
         & \# Total Params                             & 48B                             & 16B                                                  \\
         & Trained Tokens                              & 5.7T                            & 5.7T                                                           \\
        \midrule
        \multirow{5}{*}{\textit{General}}
         & RULER@128k                        & 95.4                            & -                                                       \\
         & RULER@1M                       & 94.8                            & -                                                       \\
         & GPQA-Diamond (Avg@8)                        & 71.7                            & 24.7                                                       \\
         & MMLU-Redux (EM)                             & 86.9                            & 66.9                                                  \\
         & MMLU-Pro (EM)                               & 72.7                            & 43.8                                               \\
         & FaithJudge (1-Hallu.)                       & 64.2                            &  56.0                              \\
        \midrule
        \multirow{3}{*}{\textit{Math}}
         & AIME 2025 (Avg@64)                          & 58.6                            & -                                               \\
         & MATH500 (Acc.)                             & 94.6                            & 58.0                                            \\
         & HMMT 2025 (Avg@32)                          & 44.5                            & -                                            \\
        \midrule
        \multirow{4}{*}{\textit{Code}}
         & LiveCodeBench v6 (Pass@1)                   & 45.7                   & 11.9                                          \\
         & OJBench (Pass@1)                            & 14.2                   &  -                                         \\
         & Humaneval$^+$                               & 70.9                   & 46.3                                           \\
         & MBPP$^+$                                    & 72.4                   & 56.3                                        \\
        \bottomrule
    \end{tabular}
\end{table}

\end{document}